\documentclass[12pt,journal,compsoc]{IEEEtran}

\ifCLASSOPTIONcompsoc
\else
\fi
\usepackage{graphicx}
\usepackage{cite}
\usepackage[cmex10]{amsmath}
\usepackage{amssymb}
\usepackage{epstopdf}
\usepackage{amsmath}
\usepackage{amsthm }
\usepackage{amssymb}
\usepackage{array}
\usepackage{algorithm}
\usepackage{algorithmic}
\usepackage{subfig}
\usepackage{wrapfig}
\usepackage{lipsum}
\usepackage{float}
\usepackage{subfig}
\usepackage{tikz}
\usetikzlibrary{bayesnet}

\newtheorem{theorem}{Theorem}

\ifCLASSINFOpdf
\else
\fi
\begin{document}
%
\title{Dynamic transformation of prior knowledge into\\ Bayesian models for data streams}
%
%
%
%

\author{Tran Xuan Bach, 
        Nguyen Duc Anh, 
        Ngo Van Linh, 
        and~Khoat Than
\IEEEcompsocitemizethanks{\IEEEcompsocthanksitem Tran Xuan Bach, Ngo Van Linh and Khoat Than are with Hanoi University of Science and Technology, No. 1, Dai Co Viet road, Hanoi, Vietnam.\protect\\
E-mail: tranxuanbach1412@gmail.com, linhnv@soict.hust.edu.vn and khoattq@soict.hust.edu.vn
\IEEEcompsocthanksitem Nguyen Duc Anh is with Institute for Chemical Research, Kyoto University.\protect\\
E-mail: nguyenanh.nda@gmail.com
}
\thanks{Manuscript received 5/2020.}}

\IEEEtitleabstractindextext{%
\begin{abstract}
We consider how to effectively use  prior knowledge when learning a Bayesian model from streaming environments where the data come  endlessly and sequentially. This problem is highly important in the era of data explosion and  rich sources of valuable external knowledge such as pre-trained models, ontologies, Wikipedia, etc. We show that some existing approaches can forget any knowledge very fast. We then propose a novel framework that enables to incorporate the prior knowledge of different forms into a base Bayesian model for data streams. Our framework subsumes some existing popular models for time-series/dynamic data. Extensive experiments show that our framework outperforms existing methods with a large margin. In particular, our framework can help Bayesian models generalize  well on extremely short text while other methods overfit.
An implementation of our framework is available at http://github.com/bachtranxuan/TPS.
\end{abstract}

}

\maketitle

\IEEEdisplaynontitleabstractindextext

%
\IEEEpeerreviewmaketitle

\section{Introduction}
%
%

%
%
%
%

Bayesian approach can efficiently model the uncertainty in data and make prediction on the future. A Bayesian model \cite{blei2017VIreview} however might not generalize well in the cases of \emph{misspecification} nor \emph{sparsity} nor \emph{noise}. Misspecification \cite{Box76mispecified} is a situation in which a particular model cannot cover all key aspects of reality, whereas sparsity is the case in which each data sample provides little information. Note that misspecification could not be avoided, while sparse and noisy data are prevalent in practice, such as modeling ratings or feedbacks in recommender systems \cite{huang2004applying,hidasi2016session}, and modeling short text from social networks \cite{banerjee2007clustering,yan2014btm}. Those  situations cause various challenges  \cite{huang2004applying,liang2017leveraging,ha2019eliminating,nguyen2021BPS}. Theoretically, we may not correctly recover a Bayesian model from sparse data even in cases of having arbitrarily large number of samples \cite{tang2014understandingLDA}, while in practice training from sparse and noisy data easily leads to overfitting  \cite{ha2019eliminating,nguyen2021BPS}.
One efficient way to overcome those challenges is to exploit external or prior knowledge \cite{yao2017incorporating,nguyen2015improving,zhao2017word,Ideker2011knowledge,andrzejewski2009incorporating,andrzejewski2011domain,jagarlamudi2012lexicalpriors,chen2013domain}.\footnote{Two other ways are to use multimodal data or different data sources. The latter way closely relates to \emph{Bayesian evidence synthesis} \cite{sweeting2008BayesianES,tan2018BayesianES}. This work focuses on exploitation of external knowledge, instead of data.}

We are interested in streaming environments where the data  come   sequentially and endlessly. 
\emph{How to effectively use a prior knowledge\footnote{This concept should be interpreted in a wide context, and be different with  ``prior" in the Bayesian approach where a prior is often a probability distribution.  Prior knowledge here refers to any kinds of existing knowledge that can aid a learning process.} in Bayesian models for streaming environments?} Interestingly, this question has  been rarely considered, in spite of its great significance in the era of data explosion and  rich sources of valuable prior knowledge such as pre-trained machine learning models, ontologies, Wikipedia, etc. In particular, pre-trained models have been increasingly playing a critical role in various applications \cite{devlin2019bert,Erhan2010pretrain,turian2010word}, but are mostly used in static conditions. One key reason is that streaming conditions pose various challenges, e.g., 
{How to use prior knowledge dynamically to help a Bayesian model  generalize well? Can we assure that the prior knowledge will not be forgotten quickly?}
The forgetting issue is a natural consequence of Bayes' Theorem when conditioned on large (infinite) data sets.

Some recent studies \cite{streamvb,populationdis,masegosa17power,faraji2018balancing} have provided excellent solutions to learning  Bayesian models from data streams. However, none of those methods considers exploiting external/prior knowledge. Our first contribution is to show that \emph{streaming variational Bayes} (SVB) \cite{streamvb} can forget any knowledge at a rate of $O(T^{-1})$, after learning from more $T$ minibatches of data. Such a forgetting rate in SVB is much  faster than the rate $\Omega(T^{-0.67})$ in human \cite{averell2011form}. This forgetting problem potentially appears in other related methods. As a result, those approaches cannot solve the main question of interest.

The second contribution in this paper is a novel framework called \emph{Dynamic Transformation of Prior knowledge into Bayesian models for data Streams} (TPS) that fulfils the above question and provides a unified solution to the three mentioned challenges. TPS is able to exploit knowledge which is represented by vectors, matrices, or graphs. The exploitation of prior knowledge in TPS is dynamic in nature, owing to the use of a discrete-time martingale of transformation matrices. Hence TPS helps a Bayesian model better fit with data streams and generalize on unseen observations. Finally, TPS enables us to develop a streaming learning algorithm for a base model, with few changes from an existing batch learning. This property will be beneficial in practice, since Bayesian models for static conditions are prevalent. We further show that TPS  subsumes some existing dynamic models \cite{blei2006dynamic,charlin2015dynamic} as special cases when trained on a fixed data set.

Our third contribution is an extensive evaluation of different frameworks, using two base models (\emph{latent Dirichlet allocation} (LDA) \cite{lda} for unsupervised learning, and \emph{Naive Bayes} for  classification) and three kinds of prior knowledge. The experiments show that TPS often outperforms the  state-of-the-art   methods, in terms of generalization and model interpretability \cite{lau2014machine}. In particular, TPS can help LDA and Naive Bayes generalize  well on  short text while some approaches encounter overfitting.

\textsc{Roadmap:} We first summarize closely related work. Then we present TPS and two case studies. After that we discuss some theoretical properties of TPS, and the proof about catastrophic forgetting in SVB. Extensive evaluation appears in last section and Supplement.
\hfill
\section{Related work}
There are two main directions to deal with data streams. The first direction is to design a completely new model for the endlessly sequential data \cite{blei2006dynamic,wang2006topics,wei2007dynamic,wang2008cDTM}. The other direction is to design online/streaming algorithms for learning Bayesian models, i.e., to adapt a model from static conditions to streaming ones. Efficient methods in this direction include streaming variational Bayes (SVB) \cite{streamvb}, population variational Bayes (PVB) \cite{populationdis}, online learning \cite{cappe2009onlineEM,Bottou18onlineLearning}, sequential Monte Carlo \cite{doucet2001introduction}, surprise minimization \cite{faraji2018balancing}. Interestingly, rigorous study on exploiting external/prior knowledge in streaming conditions is  rare. 

A wide range of studies have shown that an appropriate use of prior knowledge can significantly improve the model interpretability and generalization. Useful prior knowledge might be in different forms, such as similarity graphs \cite{petterson2010word,xie2015incorporating}, WordNet \cite{yao2017incorporating}, pre-trained models \cite{nguyen2015improving,zhao2017word}, or domain knowledge \cite{Ideker2011knowledge,andrzejewski2009incorporating,andrzejewski2011domain,jagarlamudi2012lexicalpriors,chen2013domain}. In particular, pre-trained models, considered as precious prior knowledge, have been  playing a crucial role in various applications \cite{devlin2019bert,Erhan2010pretrain,turian2010word}. However, most existing works just focus on non-streaming conditions.

Existing  methods  have difficulties to effectively exploit human knowledge in streaming environments. SVB learns a model by uniformly balancing the new with old knowledge learned from data, and thus only uses the external knowledge in the first step of the learning process.  This strategy can forget any knowledge very fast and limits the effect of external knowledge. (A rigorous proof can be found in Appendix B). To avoid uniformity, power priors \cite{ibrahim2015power} can be exploited to balance the old with new knowledge at each time step. One issue is that the balancing constant has to be set manually, causing a drawback in streaming conditions.   \cite{masegosa17power} remove such a drawback by considering the balancing constant as a random variable which follows a \textit{Hierarchical power prior} (HPP). Therefore, SVB-HPP  \cite{masegosa17power})  is  an elegant  combination of SVB and HPP to balance the old with new knowledge in a Bayesian way. Those observations suggest that SVB-HPP and SVB face the same difficulty when exploiting external  knowledge.
\cite{kpspakdd} suggest to maintain the prior knowledge directly in each learning step, however: the knowledge is encoded into a prior distribution which is static or gradually vanishing. Such a usage is not flexible and cannot utilize the full strength of human knowledge. Furthermore, the prior should be encoded by vectors, which therefore limits the utilization of various forms of human knowledge. In contrast, TPS in this work enables us to use richer types of knowledge, which can be represented by vectors, matrices, graphs, and pre-trained models. Further, the exploitation of knowledge in TPS is dynamic in nature.

A related topic is dynamic models for dynamic/time-series data of fixed size. Examples include \cite{blei2006dynamic,he2013dynamic,charlin2015dynamic,jahnichen18a,dieng2019dynamic}. One common limitation of most of those works is that their learning algorithms can only deal with training datasets of finite size, as many passes over the whole dataset are required in the training phase. In contrast, the learning method for TPS deals successfully with streams where the data may come sequentially and endlessly. The ability of TPS, to work with real data streams and to efficiently exploit external knowledge, goes beyond many existing dynamic models.

\section{Dynamic Transformation of Prior knowledge into Bayesian models for data Streams (TPS)}

In this section, we present the ideas of our framework. We then explicitly describe applications to LDA and Naive Bayes.

\emph{A motivating example:} 
We may want to analyze a tweet stream from Twitter to understand the hidden themes/topics ($\beta$). Each tweet contains some observed words ($x$), while each word has a hidden role ($z$) to make a meaningful tweet. The theme of the tweets can change over time, e.g., COVID-19 rarely appeared in 2019 but was frequently tweeted in 2020. One may not clearly understand about COVID-19 when first reading some tweets which are often short and noisy. In those cases, some reference knowledge ($\eta$) may facilitate his/her understandings.

\subsection{The TPS framework}
Following \cite{populationdis} and \cite{svi}, we consider a general model $B(\beta, z, x)$ with two kinds of variables: a \emph{global variable} $\beta$ of size $K \times V$ to model the latent structure that is shared among data points $x_{1:N}$, and probably a \textit{local variable} $z_{i}$ to model the latent structure that governs the $i$th data point $x_i$.\footnote{$V$ is the dimensionality of variable $x$, while $K$ represents the number of hidden factors.} Such a model is general and successfully applied in static conditions. However, there are several challenges in a streaming environment. A data stream is an \emph{infinite} sequence of minibatches $D = \{D^{1}, D^{2}, ..., D^{t},...\}$, and each minibatch $t$ consists  of $M$ observed data points: $D^{t} = \{x_{1}^t,x_{2}^t,...,x_{M}^t\}$.

Assume we have an external knowledge  $\eta$ which is represented by a matrix of size $L \times V$, where $L$ is the embedding size. Note that a matrix can help us represent different kinds of knowledge in practice, such as pre-trained word embedding \cite{mikolov2013distributed} which uses a vector to represent the meaning of a word,  the  relationships among entities, and social graphs for the connections of people. For example, the prior knowledge can come from graphs\footnote{Clearly, those graphs can be represented by adjacent matrices. \cite{bordes2013translating} further showed that we can represent any general graph knowledge into embedding spaces. The low rank matrices in the embedding spaces help to exploit the knowledge in the graph more effective.}  such as WordNet of size $V \times V$ which means $L=V$, or from word embeddings  of size $E\times V$ where $E$ is the embedding dimensionality.

In practice, the prior knowledge representations and model's variables probably have different shapes, i.e., the model parameter $\beta$ has size $K \times V$ and the prior has size $L \times V$. For this problem, we create a mapping $f$ to transform the knowledge $\eta$ into $\beta$ in each minibatch $t$.\footnote{The mapping  can be chosen as a (pre-specified) nonlinear function, a neural network,... As an example, we will use the standard {softmax} function as the mapping $f$ in the later subsections.} This  $f$ will map the linear transformation $\pi^t \eta$ into the space of $\beta$, where  $\pi^t$ is a transformation matrix of size $K \times L$. Then, the global variable $\beta^t$ at time $t$ is computed by: $\beta^t = f(\pi^t \eta)$.

There may be a dynamic of $\beta$ over time in the data stream (e.g. the theme of tweets can significantly change from 2019 to 2020). We need to model such a dynamic, and our reparameterization before translates the dynamic into $\pi$. Therefore, we make a relation between  $\pi^{t-1}$ and $\pi^t$ to capture such a dynamic. We assume  $\pi^t_k \sim \mathcal{N}(\pi^{t-1}_k, \sigma I)$, where $k$ is the row index of $\pi^t$,  $I$ is the identity matrix of size $L$, and $\sigma (\ge 0$) is the variance parameter to make $\pi^t_k$ fluctuate around $\pi^{t-1}_k$. By this way, the sequence of transformation matrices composes a discrete-time martingale. $\pi^t_k$ can also be interpreted as a Gaussian random walk. Note that $\pi^t$  plays as weighting the knowledge before transformed into the global variable of the Bayesian model. The employment of a random walk help TPS exploit the knowledge $\eta$ dynamically.

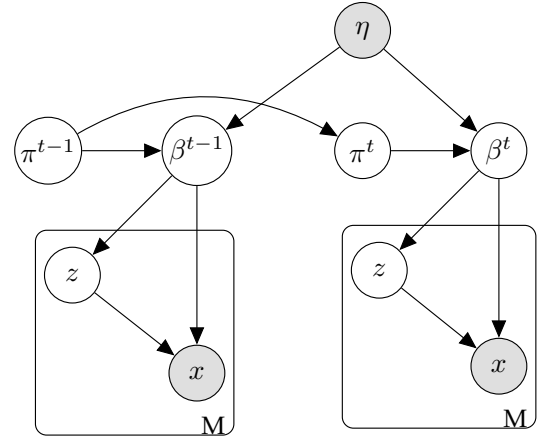
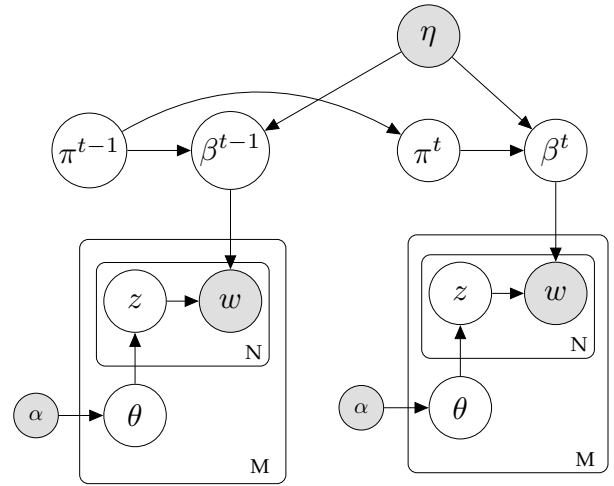
\begin{figure}[tp]
\centering
\subfloat[TPS for  $B(\beta, z, x)$ as the base model.]{
\resizebox{0.40\textwidth}{!}{
\begin{tikzpicture}
\node[latent](b){$\beta^{t-1}$};%
\node[latent, left=1 of b](p){$\pi^{t-1}$};%

\node[obs, below= 2 of b](x){$x$};%
\node[latent, below left =1 and 1 of b](z){$z$};%
\plate[inner sep = 0.1cm,yshift=0.1cm]{corpus}{(x) (z)}{M};%

\node[latent, right=3 of b](b1){$\beta^{t}$};%
\node[latent, left=1 of b1](p1){$\pi^{t}$};%
\node[obs, above = 0.8 of p1](eta) {$\eta$};%
\node[obs, below=2 of b1](x1){$x$};%
\node[latent, below left =1 and 1 of b1](z1){$z$};%
\plate[inner sep = 0.1cm, yshift=0.1cm]{corpus}{(x1) (z1)}{M};%

\edge{b}{z};%
\edge{b}{x};%
\edge{eta}{b};%
\edge{p}{b}
\edge{z}{x}

\edge{b1}{z1};%
\edge{b1}{x1};%
\edge{eta}{b1};%
\edge{p1}{b1}
\edge{z1}{x1}
\draw[->] (p) to [out =30, in =150](p1);%

\end{tikzpicture}
}
\label{fig:tpsgeneral}
}
\hfill%
\subfloat[TPS when LDA is the base model.]{
\resizebox{0.45\textwidth}{!}{
\begin{tikzpicture}
\node[latent,scale=1.4](b){$\beta^{t-1}$};%
\node[latent, left=1 of b,scale=1.4](p){$\pi^{t-1}$};%
\node[obs, below=1.25 of b,scale=1.4](w){$w$};%
\node[latent, left=0.5 of w,scale=1.4](z){$z$};%
\node[latent, below=0.8 of z,scale=1.4](theta){$\theta$};%

\node[obs, left=0.7 of theta](a){$\alpha$};%
\plate[inner sep=0.1cm,yshift=0cm]{document}{(z) (w)}{N};%
\plate[inner sep = 0.25cm,yshift=0.1cm]{corpus}{(theta) (document)}{M};%

\node[latent, right=4 of b,scale=1.4](b1){$\beta^{t}$};%
\node[latent, left=1 of b1,scale=1.4](p1){$\pi^{t}$};%
\node[obs, above = 0.8 of p1,scale=1.4](eta) {$\eta$};%

\node[obs, below=1.25 of b1,scale=1.4](w1){$w$};%
\node[latent, left=0.5 of w1,scale=1.4](z1){$z$};%
\node[latent, below=0.8 of z1,scale=1.4](theta1){$\theta$};%
\node[obs, left=0.7 of theta1](a1){$\alpha$};%
\plate[inner sep=0.1cm,yshift=0cm]{document1}{(z1) (w1)}{N};%
\plate[inner sep = 0.2cm,yshift=0.1cm]{corpus1}{(theta1) (document1)}{M};%

\edge{a} {theta};%
\edge{theta} {z};%
\edge{z}{w};%
\edge{p}{b};%
\edge{b}{w};%
\edge{eta}{b};%

\edge{a1} { theta1};%
\edge{theta1} {z1};%
\edge{z1}{w1};%
\edge{p1}{b1};%
\edge{b1}{w1};%
\edge{eta}{b1};%
\draw[->] (p) to [out =30, in =150](p1);%

\end{tikzpicture}
}
\label{fig:tpslda}
}

\caption{Graphical representation for TPS. TPS will exploit external knowledge $\eta$ and move a base model ($B(\beta, z, x)$ in (a) or LDA in (b)) through time, owing to the transformation $\pi$.}
\label{fig:GTPS}
 \vspace{-10pt}
\end{figure}

Given the global variable $\beta^t$ in each minibatch $t$, the generative model of data points is the same as those in the original  $B$. The graphical representation of TPS appears in Figure \ref{fig:tpsgeneral}.

\textbf{Learning in TPS:} 
When facing with sequential data, many approaches \cite{doucet2001introduction} often formulate the learning  as the Bayesian filtering problem for which one has to estimate the posterior $p(\pi^1, \pi^2, ..., \pi^t | D^1,D^2,...,D^t)$ or $p(\pi^t | D^1,D^2,...,D^t)$. Note that estimating one of those posteriors will require all past data, and thus is impractical for data streams, as $t \rightarrow \infty$. 
Here we propose an entirely different approach which avoids reusing past data. The learning process is performed in each minibatch $t$ by maximizing the posterior  $p(z,\pi^t | \pi^{t-1}, \eta ,D^t)$, where $\pi^{t-1}$ is made available from the previous minibatch. Hence, our approach will be potentially more efficient and truly applicable to data streams. We will decompose the posterior into components in order to reuse the inference steps of the original model $B$ as:
$ p(z,\pi^t|\pi^{t-1},\eta,D^t) \propto p(z,\pi^t,D^t | \pi^{t-1}, \eta) \propto p(\pi^t |\pi^{t-1})p(z,D^t |  \eta,\pi^t) \propto p(\pi^t | \pi^{t-1})p(z,D^t | \beta^t). 
$
 
 In $\log$ form, we have:
 
 \begin{align}
 \mathrm{LP}(z,\pi^t) =& \log p(z,\pi^t|\pi^{t-1}, \eta ,D^t)\nonumber \\
          = & \log  p(\pi^t|\pi^{t-1}) + \log p(z,D^t|\beta^t) + \mathrm{const.} \label{eq:logpost_gtps}
 \end{align}
 
The learning process is separated into two parts for local and global variables, respectively. While the inference of local variables $(z,x)$ is inherited from the original model $B$ (e.g., by maximizing or sampling from  $p(z,x|\beta^t)$), we focus on maximizing $\mathrm{LP}$ w.r.t. $\pi^t$. We extract the component $G(\beta^t) = G(f(\pi^t \eta))$, that contains $\beta^t$, from $\log p(D^t,z |\beta^t)$. Then, we obtain the objective function $\mathrm{LP}(\pi^t) =  \log  p(\pi^t | \pi^{t-1}) + G(f(\pi^t \eta))$, and maximize it by using gradient ascent. Algorithm \ref{algo:G_TPS} briefly describes the learning process.
\vspace{-10pt}
 
\begin{table} 
\begin{minipage}{0.48\textwidth}
\begin{algorithm}[H]
    \caption{Learning in TPS}
	\label{algo:G_TPS}
\begin{algorithmic}
\REQUIRE{ Prior knowledge $\eta$, mapping $f$, variance $\sigma$, data sequence $\{D^1,D^2,...$\}}
\ENSURE{ $\pi$ \\}
Initialize  $\pi^0 $ randomly\\
\FOR {minibatch $t = 0,1,...$} 
\STATE{Receive a minibatch $D^t$ of data}
\WHILE{not convergence}
\STATE {\emph{Do inference} w.r.t. the local variables $(z,x)$, given $\beta^t = f(\pi^{t} \eta)$ and $D^t$} \\ \hspace{10pt} (e.g., by maximizing or sampling from  $p(z,x|\beta^t)$)\\
\STATE{\emph{Maximize} (\ref{eq:logpost_gtps}) w.r.t $\pi^t$, given the statistics from $(z,x)$}
\ENDWHILE
\STATE{Set $\pi^{t+1}:=\pi^t$}
\ENDFOR
\end{algorithmic}
\end{algorithm} 
\end{minipage} 
\begin{minipage}{0.48\textwidth}
\begin{algorithm}[H]
    \caption{TPS training for LDA}
	\label{algo:TPS_LDA}
\begin{algorithmic}
\REQUIRE{ Prior knowledge $\eta$, hyper-parameter $\alpha$, variance $\sigma$, data sequence $\{D^1,D^2,...\}$}
\ENSURE{$\pi$ \\}
Initialize $\pi^0$ randomly\\
\FOR  {minibatch $t = 0,1,...$} 
\STATE{Receive a minibatch $D^t$ of data}
\WHILE{not convergence}
\STATE {\emph{Infer} ($\gamma_d, \phi_d$) for each document $d \in D^t$ by iteratively computing (\ref{eq:phi}) until convergence, given $\beta_k = \mathrm{softmax}({\pi_k^t} {\eta})$ for each $k$}
\STATE{\emph{Maximize} (\ref{Eq:posterior_pi}) w.r.t $\pi^t$} 
\ENDWHILE
\STATE{Set $\pi^{t+1}:=\pi^t$}
\ENDFOR
\end{algorithmic}
\end{algorithm}
\end{minipage}
\end{table}

\subsection{Case study 1: TPS when LDA is the base unsupervised model}

Next we discuss how to apply TPS to LDA \cite{lda}, one of the most popular Bayesian models. LDA consists of two global variables $(\beta, \alpha)$:  $\alpha$ contributes to the  topic mixture $\theta$ of each document and is fixed in this case study, and $\beta = (\beta_1, \beta_2, ..., \beta_K)$ where each $\beta_k$ is the topic distribution  over $V$ words.

Suppose that there is an available prior knowledge $\eta$ of size $L \times V$. We incorporate the prior knowledge into $\beta$ by a  linear transformation with a transformation matrix $\pi$ of size $K\times L$, and then followed by the {softmax} operator. 
The generative process of the documents in  minibatch $t^{th}$ is as follows (Figure~\ref{fig:tpslda}):
\begin{enumerate}
\item Draw the transformation matrix: \\ $\pi_k^t  \sim \mathcal{N}(\pi_k^{t-1},\sigma^2 I)$
\item Calculate the topic distributions: \\ 
    $\beta_{k} = \mathrm{softmax}({\pi_k^t} {\eta}), k \in [K]$
\item For each document $d$ of length $N_d$:
\begin{enumerate}
\item Draw a mixture: $\theta_d  \sim \mathrm{Dirichlet}( \alpha )$
\item For the $ i^{th} $ word of $d$:
Draw topic index $z_i  \sim \mathrm{Multinomial}(\theta_d)$
and then draw word  $w_i  \sim \mathrm{Multinomial}(\beta_{z_i})$
\end{enumerate} 
\end{enumerate}

\textbf {Learning parameters:}  We apply Algorithm \ref{algo:G_TPS} for estimating the posterior. We emphasize that our framework utilizes the available inference methods (e.g., variational inference,  Gibbs sampling) for local variables $(w,\theta,z)$ in the original LDA model.

Here, we use mean-field variational inference as in the original paper \cite{lda}:
$q(\theta_d,z_d | \gamma_d, \phi_d) = {q(\theta_d | \gamma)}{\prod_{n=1}^{N_d}{q(z_{dn}|\phi_{dn})}} $
with the variational distributions: $q(\theta_d|\gamma_d) = \mathrm{Dirichlet}(\gamma_d)$ and $q(z_{dn}|\phi_{dn}) = \mathrm{Multinomial}(\phi_{dn})$ where $\gamma_d$ and $\phi_d$ are variational parameters w.r.t.  document $d$. According to \cite{lda}, the inference for document $d$ reduces to repeating the following updates until convergence:
\begin{gather}
\gamma_{dk} = \alpha_k + \sum_{n \in [N_d]}\phi_{dnk} \nonumber \\
\phi_{dnk} \propto \exp \psi(\gamma_{dk}) \cdot \exp (\sum_{v \in [V]} I[w_{dn}=v] \log \beta_{kv})\label{eq:phi}
\end{gather}
where $[V] = \{1, ..., V\}$, $I$ is the indicator function, $\psi$ is the digamma function, $k \in [K]$.

The component depending on the global variable $\pi_k^t$ in (\ref{eq:logpost_gtps}) for each $k$ given data $D^t$  is:
$
\mathrm{LP}(\pi_k^t) = \sum_{d \in D^t} \sum_{n \in [N_d]} \log p(w_{dn}|z_{dn},\beta) + \log  p(\pi_k^t | \pi_k^{t-1}) 
        = - \frac{1}{2\sigma} \parallel \pi^t_k -\pi^{t-1}_k \parallel^2_2   +\sum_{d \in D^t} \sum_{n \in [N_d], v \in [V]} \phi_{dnk}I[w_{dn}=v]\log \beta_{kv}
$, after removing some constants.
In more details, 
\begin{align}
&\mathrm{LP}(\pi_k^t) = - \frac{1}{2\sigma} \parallel \pi^t_k -\pi^{t-1}_k \parallel^2_2  \nonumber \\
            &+ \sum_{d \in D^t} \sum_{n,v}^{N_d, V} \phi_{dnk}I[w_{dn}=v] \pi_k^t \eta_v   \nonumber \\
            &-\sum_{d \in D^t} \sum_{n,v}^{N_d, V} \phi_{dnk}I[w_{dn}=v] \log  {\sum_{i \in [V]} \exp ({\pi_k^t} {\eta_i}) }
\label{Eq:posterior_pi}
\end{align}
Consider the concavity of function $\mathrm{LP}(\pi_k^t)$. 
It is obvious that $- \frac{1}{2\sigma} \parallel \pi^t_k -\pi^{t-1}_k \parallel^2_2 $ and $ {\pi_k^t} \eta_v $ are concave functions with respect to $\pi_k^t$. Further, the log-sum-exp function is well-known convex. Therefore, $\mathrm{LP}(\pi_k^t)$ is concave with respect to $\pi_k^t$, and we can use gradient ascent to find its maximum. We can sum up the learning algorithm of TPS for LDA as in Algorithm \ref{algo:TPS_LDA}.

\subsection{Case study 2: TPS when Naive Bayes is the base supervised model}

In this subsection, we apply TPS to Multinomial Naive Bayes for  classification on document streams. Let $C$ be the number of classes, $\beta_c$ be the class distribution over $V$ words of the vocabulary (where $\beta_{cj} = P(j|c)$ and $\sum_{j \in [V]} \beta_{cj}=1$) for each $c \in [C]$. Each document $d$ belonging to class (label) $c_d$  is represented by a bag of $N_d$ words and each word $w_{d,i}$ is generated from $\mathrm{Multinomial}(\beta_{c_d}) $.

Suppose that we have a prior knowledge  $\eta$ of size ${L \times V}$. The generative process of documents in the minibatch $t^{th}$ is as follows: For each class $c$, draw   $ \pi_c^t \sim \mathcal{N}(\pi_c^{t-1},\sigma^2 I)$ and calculate $\beta_{c} = \mathrm{softmax}({\pi_c^t} {\eta})$. Generate  document $d$ by  drawing class label $c_d  \sim \mathrm{Multinomial}(\alpha)$ and then drawing each word $w_{dn}  \sim \mathrm{Multinomial}(\beta_{c_d})$.

\textit{Learning:} From (\ref{eq:logpost_gtps}), we extract the term associated with $\pi_c^t$ for each class $c$ as:
{\small 
\begin{align*}
&\mathrm{LP}(\pi_c^t) = \log  p(\pi_c^t | \pi_c^{t-1}) + \sum_{d \in D_c^t} \sum_{n \in [N_d]} \log p(w_{dn}|c_{d},\beta) 
\end{align*}
\begin{align*}
 =& - \frac{1}{2\sigma} \parallel \pi^t_c -\pi^{t-1}_c \parallel^2_2 + \sum_{d \in D_c^t} \sum_{n \in [N_d]} \sum_{v \in [V]} I[w_{dn}=v]\log \beta_{cv}  \\
 =& - \frac{1}{2\sigma} \parallel \pi^t_c -\pi^{t-1}_c \parallel^2_2  + \sum_{d \in D_c^t} \sum_{n \in [N_d]} \sum_{v \in [V]} I[w_{dn}=v] ({\pi_c^t} {\eta_v})  \\
 & -\sum_{d \in D_c^t} \sum_{n \in [N_d]} \sum_{v \in [V]} I[w_{dn}=v] \log  {\sum_{v \in [V]} \exp({\pi_c^t} {\eta_i})}    
\end{align*} }
where $D_c^t$ denotes the documents with class label $c$ in minibatch $t$.
Learning for NB is really simple. At each minibatch $t$, we use gradient ascent to maximize $\mathrm{LP}(\pi_c^t)$ with respect to $\pi_c^t$, for each  $c$. $\alpha = \frac{1}{C}$ is used in our experiments. 

\section{Some properties of TPS} 
TPS has several advantages. Firstly, TPS can exploit different forms of prior knowledge such as vectors,  graphs, and matrices. 
Thanks to the mapping $f$, TPS can transform the prior knowledge into the desired size of the global variable. Existing methods, e.g. SVB, PVB, SVB-HPP, are limited in this aspect. Secondly, TPS enables a base model, designed for static conditions, to work well in a streaming environment.  

Thirdly, when trained from a dataset of bounded size, TPS subsumes many existing dynamic models \cite{blei2006dynamic,he2013dynamic,charlin2015dynamic}. For example, when the prior $\eta$ is the identity matrix of size $V \times V$ and LDA is the base model,  TPS is reduced to dynamic topic models  \cite{blei2006dynamic}. It is worth noting that the learning algorithms for those dynamic models can work with only datasets of fixed size, whereas the learning method for TPS deals successfully with streams with infinite size. The ability of TPS to work with real data streams and to efficiently exploit external knowledge  is a significant advantage.

Next, we will analyze two key properties.


\subsection{Balancing the old, new, and external knowledge}
The ability to balance the old and new knowledge is the basic requirement for a learning system. When learning from data streams, three main sources of knowledge should be considered: the old knowledge learned in past data, the new knowledge to be learned from incoming data, and the external knowledge. TPS has a simple mechanism to balance those three sources, owing to the objective function in (\ref{eq:logpost_gtps}):
\begin{align*}
\mathrm{LP}(z,\pi^t_k) =& - \frac{1}{2\sigma} \parallel \pi^t_k -\pi^{t-1}_k \parallel^2_2 
+ \log p(z,D^t|\beta^t)\\
&+ \mathrm{const}.
\end{align*}
The first term controls the flexibility of the new model. An increase in variance $\sigma$ implies that the new model at time $t$ might be far from the previous one, and thus the new model is searched in a larger region. As $\sigma \rightarrow \infty$, TPS will not remember what have been learned before. In contrast, a decrease in $\sigma$ implies the new model should not be far from the previous one. As $\sigma=0$, we cannot learn any new knowledge at all since the first term dominates $\mathrm{LP}(z,\pi^t_k)$.

The second term, $\log p(z,D^t|\beta^t)$, enables TPS to learn new knowledge from new data. Different with the static use of external knowledge in KPS \cite{kpspakdd}, TPS exploits the prior dynamically owing to the use of the transformation matrix $\pi^t$. Estimation of $\pi^t$ at each minibatch implies the dynamic balancing between the prior and the new knowledge learned from the data at time $t$. Note that the variance $\sigma$ also plays the key role in this balance: lower $\sigma$ means less knowledge can be learned from new data. From those observations, one can see that TPS provides a simple mechanism ($\sigma$) to dynamically balance three sources of knowledge, overcoming the limitation of existing methods.

\subsection{Catastrophic forgetting}
A serious issue in many learning methods is catastrophic forgetting \cite{parisi2019continual}, i.e., the learned knowledge can be forgotten quickly as learning from more data/tasks. This issue has been found repeatedly for neural networks, but was unclear for Bayesian models. More importantly, existing works did not theoretically show how fast a method can forget. Here, we show that SVB \cite{streamvb} has a fast forgetting rate. The detailed proof appears in Appendix B.

\begin{theorem}[Forgetting in SVB for LDA]
Let $\xi^0$ be the model at time 0, and $\xi^t$ be the model after learning by SVB from more $t$ minibatches. Then 
$
\| \xi^t - \xi^0 \|_1 \ge t$ and 
${\|\xi^0 \|_1}  = O(t^{-1}) \cdot {\| \xi^t \|_1},
$
suggesting that $\xi^0$ will be quickly forgotten, at a rate of $O(t^{-1})$, in the learned model $\xi^t$.
\end{theorem} 

It can be shown that this property of SVB holds for Naive Bayes and  a large class of LDA-based variants which are conjugate. Such a forgetting rate in SVB is much  faster than the rate $\Theta(t^{-0.67})$ in human \cite{averell2011form}. We conjecture that  a fast rate might appear in many existing methods. In contrast, TPS does not encounter this problem. It has an explicit mechanism to balance the three sources of knowledge as discussed in the last subsection. By manipulating $\sigma$, TPS can remember the knowledge better.

\section{Experimental evaluation}

In this section, we conduct extensive experiments to evaluate the performance of TPS. Further quantitative and qualitative evaluations can be found in the appendices.

\subsection{Unsupervised learning for LDA} \label{sec-unsupervised-LDA}

We first evaluate TPS when applied to LDA. We take four state-of-the-art baselines: \textbf{SVB} \cite{streamvb}, \textbf{PVB} \cite{populationdis}, \textbf{SVB-PP} \cite{masegosa17power}, and \textbf{KPS} \cite{kpspakdd}.\footnote{SVB-HPP  is not included since its application to LDA requires non-trivial efforts. Further, as observed by \cite{masegosa17power}, SVB-HPP is often comparable to the best SVB-PP. \\ 
Except KPS, all of SVB, PVB, and SVB-PP do not explicitly exploit external/human knowledge and can only use the prior $(\eta)$ at the initialization. Therefore, for a fair comparison, we encode the external knowledge in the initialization of those baselines. Whenever the forms of prior knowledge are unsuitable for the baselines, we use PCA to transform the edge matrices to the same shape with $\eta$.}

\textbf{Datasets:} We use 2 regular text (Grolier, TMN)  and 4 short text datasets with some statistics in Table \ref{data}.\footnote{\emph{Grolier} from http://cs.nyu.edu/$\sim$roweis/data.html, \emph{TMN} from http://acube.di.unipi.it/tmn-dataset/,  \emph{NYT-title} from http://archive.ics.uci.edu/ml/datasets/Bag+of+Words/; \emph{Yahoo-title, TagMyNews-title (TMN-title), Irishtimes} from http://www.kaggle.com/therohk/ireland-historical-news/} Those short text corpora contain documents of extremely short length, and are used in our evaluation to help us see the role of prior knowledge in the cases of extreme sparsity.

\begin{table}
\caption{Some statistics about the datasets. For Irishtimes, we use documents of the next minibatch to evaluate the model at any minibatch.}
\label{data}
\begin{center}
\begin{tabular}{lrrrr}
\hline
Dataset & Vocabulary  & Training  & Testing  & words \\
 &  size &  size &  size &per doc \\ \hline
Grolier   & 15,269 & 23,044 & 1,000 & 79.9 \\ 
TMN & 11,599 & 31,604 & 1,000 & 24.3 \\ 
NYT-title & 46,854 & 1,664,127 & 10,000 & 5.0 \\ 
Yahoo-title & 21,439 & 517,770 & 10,000 & 4.6 \\ 
TMN-title & 2,823 & 26,251 & 1,000 & 4.6 \\ 
Irishtimes & 28,816 & 1,374,669 & - & 5.0 \\ \hline
\end{tabular}
\end{center}
\end{table} 

\textbf{Prior knowledge:} We use \emph{word embedding} and \emph{word graph} as two kinds of prior knowledge. The word embeddings were pre-trained  from 6 billion tokens of Wikipedia2014 and Gigaword5 by \cite{pennington2014glove}\footnote{http://nlp.stanford.edu/projects/glove/}. Each word is represented by a $200$-dimensional vector ($L=200$).
\\Word graph represents the relationships among words, and is represented by a matrix of size $V \times V$. We build the $500$-nearest neighbor graph based on the cosine similarity of word embedding vectors, and utilize it as prior knowledge. Due to the high computational cost as working with a matrix of size $V \times V$, we only did experiments on Grolier and TMN-title.


\textbf{Evaluation metrics:} \emph{Log predictive probability} (LPP) \cite{svi} and \emph{Normalized pointwise mutual information} (NPMI) \cite{lau2014machine} are used.  While LPP measures the generalization of a  model on unseen data, NPMI examines the coherence and interpretability of the learned topics. Details about how to compute those quantities can be found in Supplement.

\textbf{Settings:} We simulate streaming data by dividing a dataset into a sequence of minibatches with batchsize: 500 for \{Grolier, TMN, TMN-title\}, 5000  for \{NYT-title, Yahoo-title\}. For LDA, we set $\alpha=0.01$, $K=50$ topics for \{Grolier, TMN, TMN-title, Irishtimes\} and  $K=100$  for \{NYT-title, Yahoo-title\}. 
We use a grid search to select suitable hyperparameters for the baselines, and report the best parameter sets for each method and each dataset. The ranges of the parameters are:   multiple power prior $\rho \in \{0.6, 0.7, 0.8, 0.9, 0.99\}$ for SVB-PP,   population size in $\{10^2, 10^3, 10^4, 10^5, 10^6\}$ for PVB,   dimming factor $\kappa \in \{0, 0.01, 0.03, 0.07, 0.1, 0.6, 0.7, 0.8, 0.9 \}$ in KPS, and  $\sigma \in \{0.01, 0.1, 1, 10, 100\}$ for TPS.

\begin{figure*}[!t]
\subfloat[Predictive probability]{\includegraphics[width=0.5\textwidth]{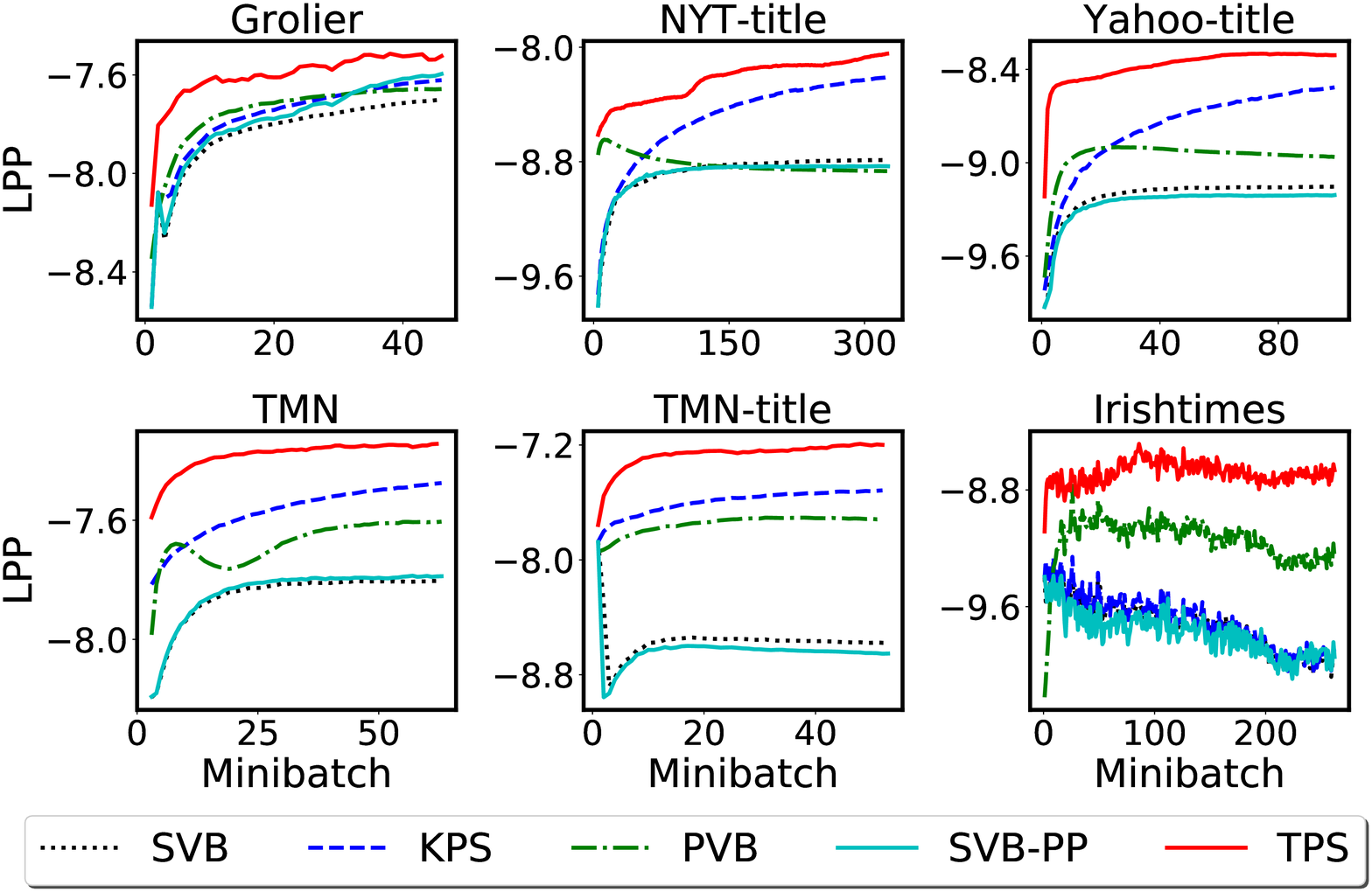} \label{fig:perplexities}} 
\subfloat[NPMI]{\includegraphics[width=0.5\textwidth]{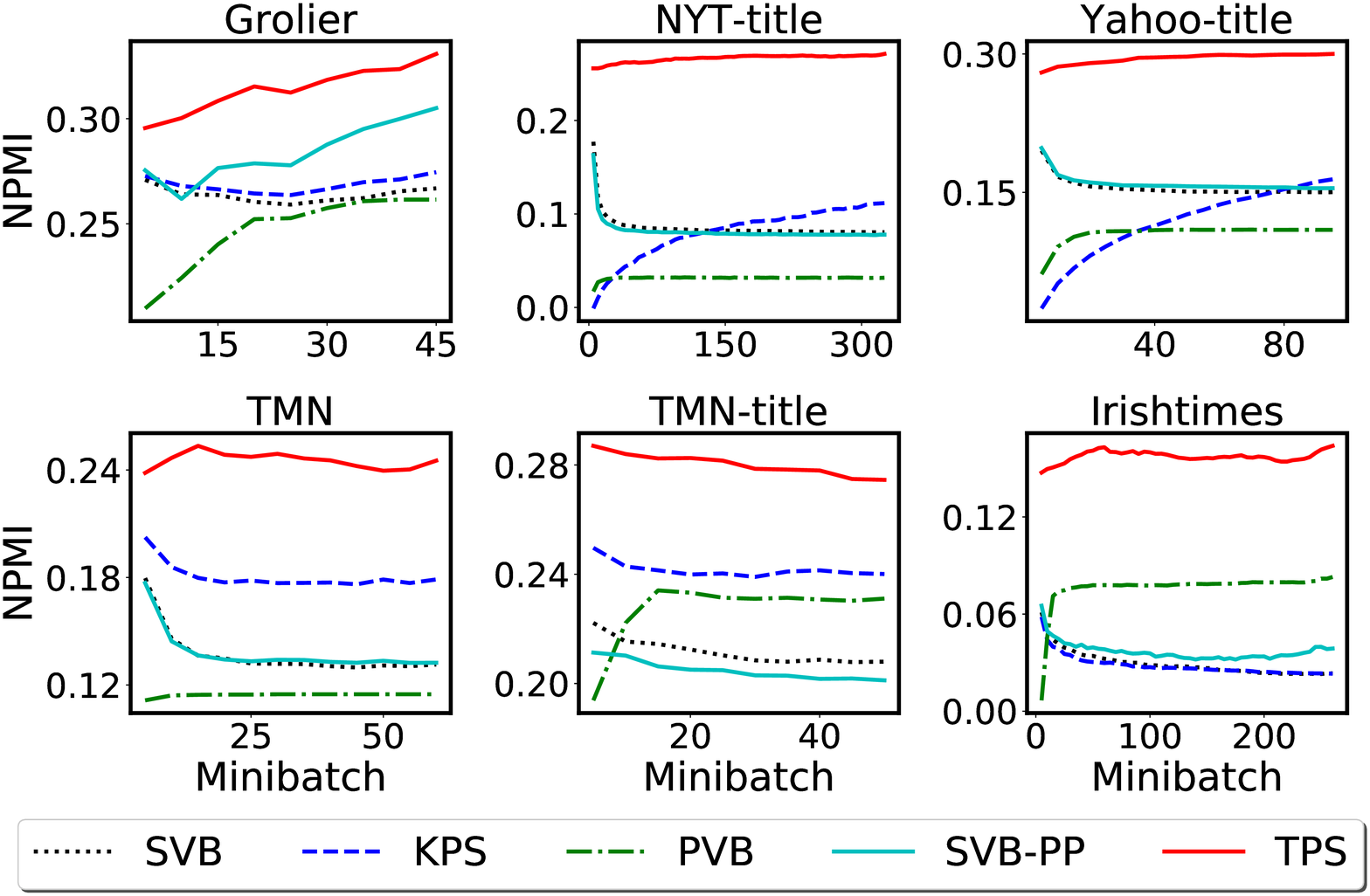} \label{fig:npmi}}

\caption{Performance of five methods when \textit{pre-trained word embeddings} is the prior knowledge and LDA is the base model. Higher is better.}
\vspace{-10pt}
\end{figure*}

\textbf{Results:}

\begin{figure*}[!t]
\centering
\includegraphics[width=0.8\textwidth]{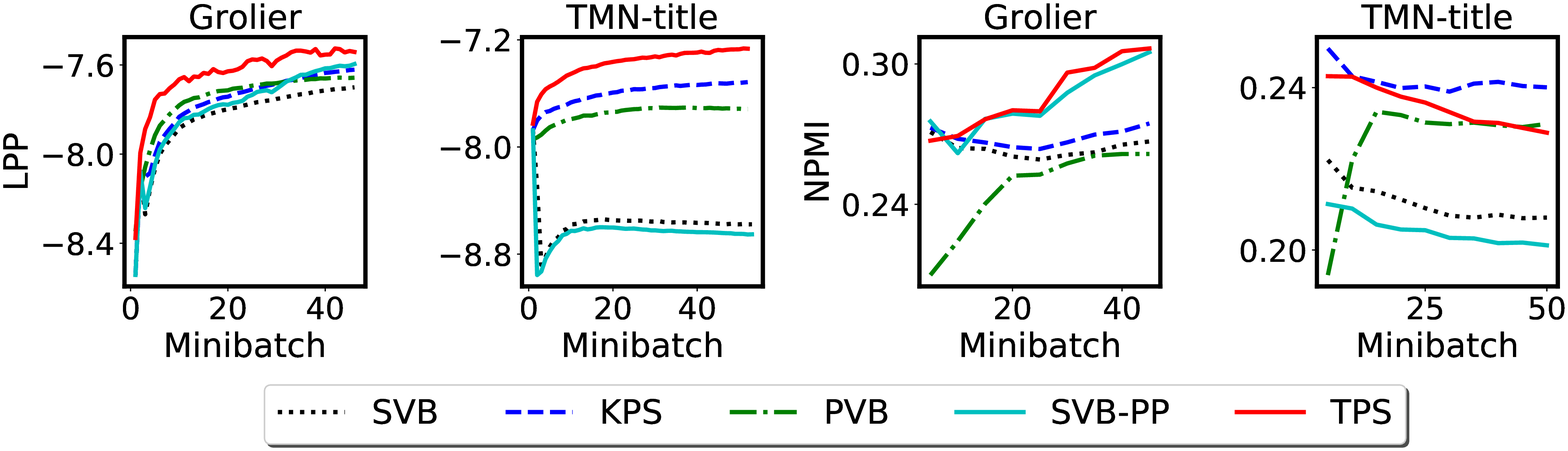}
\caption{Performance  when \textit{word graph} is used as prior knowledge, and  LDA is the base model.}
\label{fig:wn-npmi-lpp}
\end{figure*}

\textit{Predictive capacity:} Figure \ref{fig:perplexities}  and Figure \ref{fig:wn-npmi-lpp} show the results when using word embedding and word graph priors respectively. It is obvious that TPS  with both kinds of prior performs significantly better than the baselines, often by a large margin. In particular, thanks to the dynamic use of prior knowledge  in each minibatch, TPS keeps increasing the predictive ability when receiving more data. Moreover, TPS can attain very high predictive capacity from some beginning stages of the learning process. For regular text data, the predictive ability in the beginning minibatch is extremely higher than the baselines. This suggests that the knowledge from the prior contains a large amount of information, and TPS can exploit the knowledge better than KPS. 

It is worth noticing that  SVB and SVB-PP seem not to work well with extremely short text, since their predictive capability decreases as learning from more data. Short text often  does not provide enough information and clear context \cite{hong2010empirical,yan2013biterm}, and hence cause various difficulties for SVB, SVB-PP, PVB,  and KPS. KPS is able to use prior knowledge, however its ability seems to be limited because its usage of the knowledge is static along the learning process. Figure \ref{fig:perplexities}  and Figure \ref{fig:wn-npmi-lpp} clearly demonstrate that existing methods are prone to overfitting on short text, whereas TPS generalizes well.

\textit{Topic coherence:} The results of evaluating topic coherence using NPMI are reported in Figures~\ref{fig:npmi} and \ref{fig:wn-npmi-lpp}. 
With word embedding prior, TPS obtains  the best results often with a large margin. Again, TPS is effective for short text. The information from the prior injects the knowledge of word's relationship to the model. For using word graph prior, Figure~\ref{fig:wn-npmi-lpp} shows that TPS is stable in the best methods. Interestingly, KPS performed better than TPS for TMN-title. It seems that TPS did not exploit the full advantage of this knowledge, although its predictiveness is still the best.

\subsection{Balancing and sensitivity analysis} \label{sec-sensitivity}

\begin{figure}[!t]
\subfloat[Effectiveness of prior and transition model.]{
\includegraphics[width=0.5\textwidth]{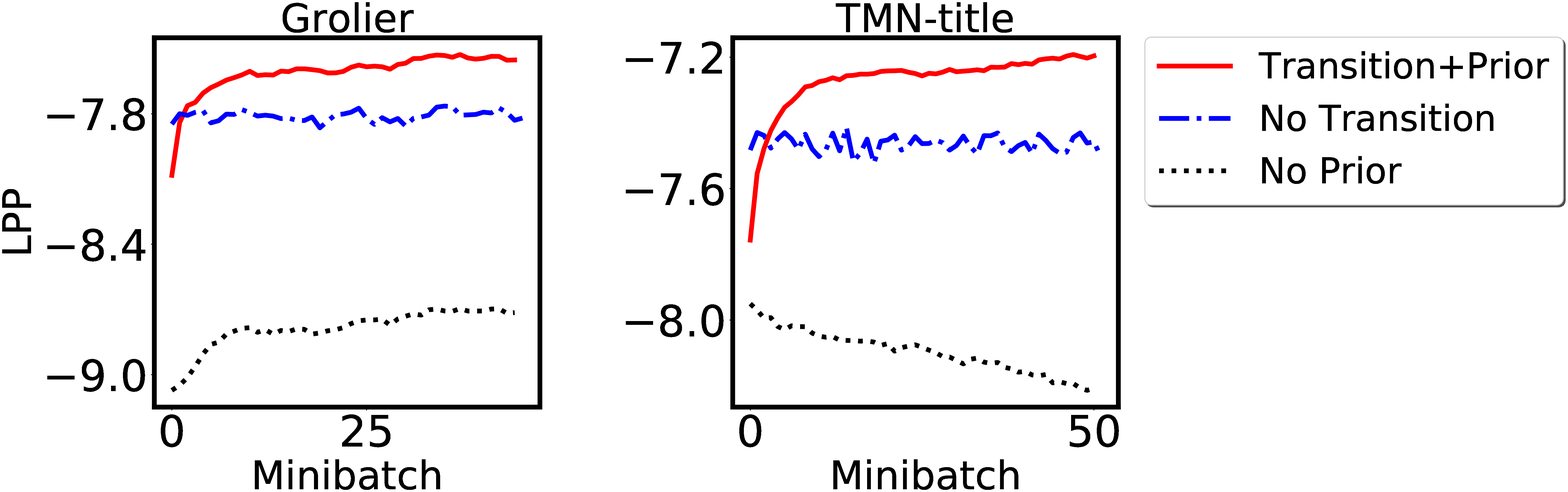} \label{fig:prior}}\\
\subfloat[Sensitivity of  $\sigma$.]{\includegraphics[width=0.45\textwidth]{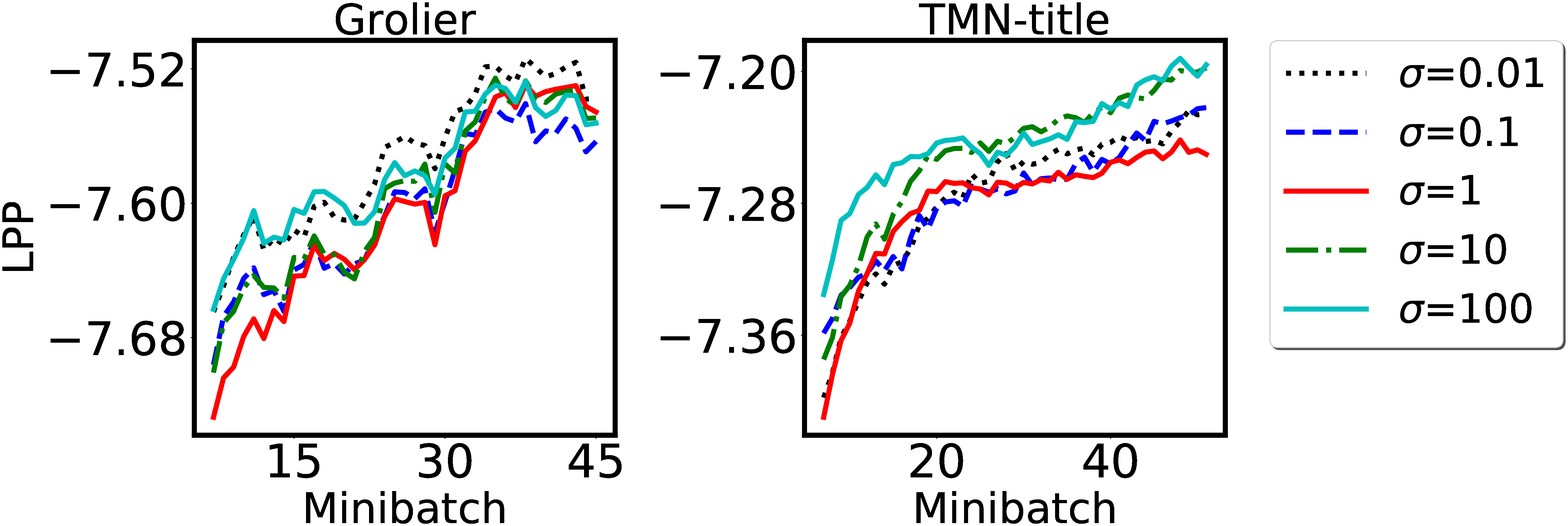} \label{fig:sigma}}
\caption{Sensitivity of TPS with respect to the key components. LDA is used as the base model. }
\vspace{-10pt}
\end{figure}

\textit{The role of prior knowledge and transition model:}
There are two important components which can significantly affect the performance of TPS: the prior knowledge $\eta$, and the transition model ($\pi^t_k \sim \mathcal{N}(\pi^{t-1}_k, \sigma I)$) which connects the  models in two consecutive time steps. We would like to see which one is really important to the performance of TPS. To this end, we take LDA as the base model,  fix batchsize = 500,  $\sigma=1, K=100$, and pre-trained word embedding as prior.  Figure \ref{fig:prior} shows the performance of TPS in three versions.  One can observe that when there is no prior, TPS does not  perform well and even encounters overfitting in short text. When a good prior knowledge is available, TPS performs significantly better and do not encounter overfitting. The transition model plays a good role as removing it may result in worse performance. It is worth observing that TPS tends to be better as learning from more data. This suggests that the prior knowledge does not overwhelm the data, but supports TPS to learn better.

\textit{Sensitivity of  $\sigma$: } Grolier (regular text) and TMN-title (short text) are used in this evaluation. We  fix the batchsize to $1000$ and $K=100$ topics. The results are presented in Figure \ref{fig:sigma}. This figure shows that one should use small $\sigma$ for long text, and large $\sigma$ for short text. The reason might be that short text contains little information and few changes will likely lead to a great variance in the meaning of that text. Therefore, the new model $\pi^{t}$ should be learned in a large region around $\pi^{t-1}$ to capture large variance in incoming data. This coincides well with our theoretical analysis.

\subsection{Streaming classification with Naive Bayes}
We compare TPS with SVB and KPS when applied to Naive Bayes for streaming classification. We  use grid search to find the best $\kappa$ in KPS. For TPS, we use $\sigma = 1$.

\begin{figure}[!t]
\centering
\hspace{-10pt}
\includegraphics[width=0.49\textwidth]{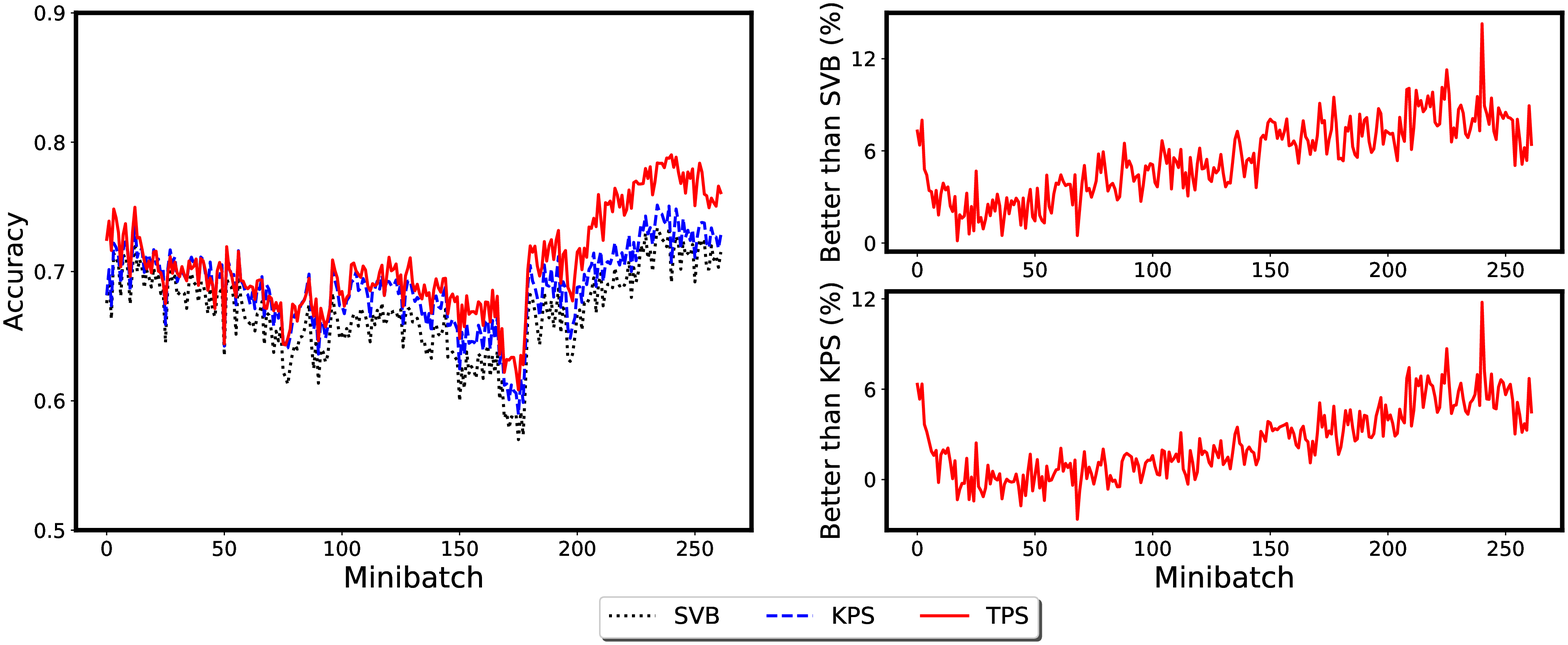}
\caption{Classification accuracy of three methods. The first subfigure shows the accuracy, while the other two subfigures show the relative improvement of TPS over   SVB and KPS, respectively. The improvement of TPS over method $A$ is measured by ${(TPS - A)}/{A}$.}
\label{fig:classification}
\vspace{-10pt}
\end{figure}

\textbf{Dataset:} We use Irishtimes which consists of $6$ categories (\textit{business, culture, news, opinion, sport, and letters}). We continuously train a model when a minibatch arrives, then do classification for documents in the next minibatch. Here, each minibatch contains the documents of a month.

\textbf{Prior knowledge:} We  extract a feature $V$-dimensional vector of each class $c$ whose element $j$ is the ratio of the number of word $j$ appeared in the class $c$ to the number of documents containing word $j$. Then, we gain a matrix $C \times V$ in which each term $v$ is represented by a $C$-dimensional vector. This matrix is used as prior for SVB and KPS. In TPS, we identify each word $v$ by concatenating a one-hot vector ($V$-dimension) and the $C$-dimensional vector in order to get a sufficient representation. We use this representation as prior knowledge.   


\textbf{Results:} Figure \ref{fig:classification} reports the accuracies of three methods. TPS is comparable to KPS in the first $100$ minibatches, better about $1-8.3\%$ than KPS in the  remaining minibatches. We observe that the prior knowledge is definitely suitable for KPS as it helps KPS to obtain high accuracy. The gap between TPS and KPS is  significant when the number of  minibatches is large. In contrast, SVB only utilizes the knowledge at the first step, and hence often gets lower accuracy than the other methods. Note that TPS performs significantly better than both KPS and SVB in the last 100 minibatches. It is worth noting that at some sudden changes in the data distribution, the performance of SVB and KPS drops significantly. TPS can reduce such a bad effect of those sudden changes. The main reason may come from the effective exploitation of prior knowledge. This seems to be an advantage of TPS in changing environments.


\subsection{Utilization of the full strength of the original knowledge}

\begin{figure}[!t]
\includegraphics[width=0.5\textwidth]{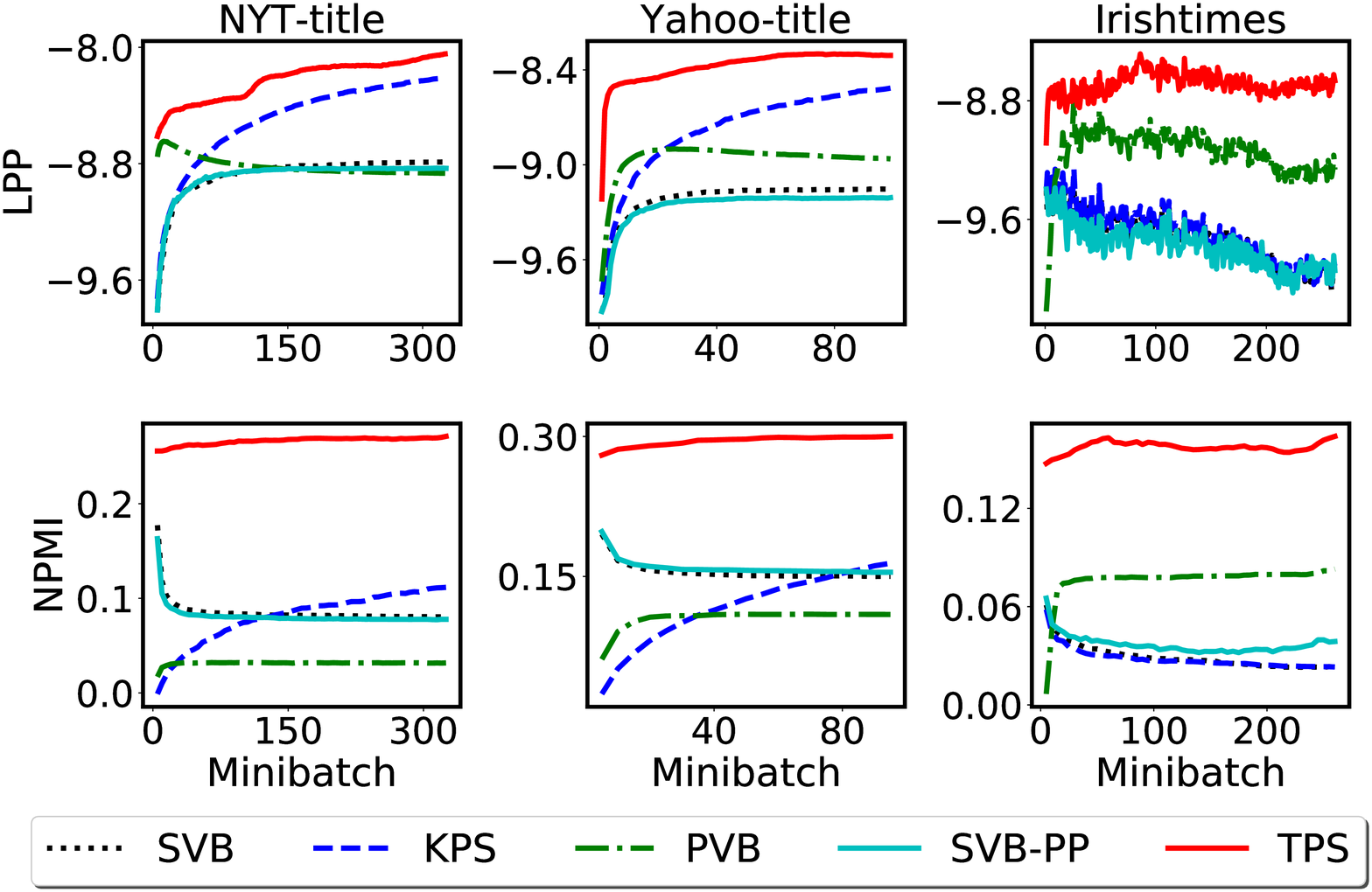} 
\caption{Performance of five methods when \textit{pre-trained word embeddings} is in its original representation.}
\label{fig:LPP-orig}
\vspace{-10pt}
\end{figure}

The final evaluation is to see how well can the baselines utilize the full strength of external knowledge. The experiments with Naive Bayes in the previous subsection provide some good evidences as all methods can use the original knowledge. However, in the experiments with LDA in subsection \ref{sec-unsupervised-LDA} we have to transform the knowledge (pre-trained word embedding) into a form that can be used in SVB, SVB-PP, PVB, and KPS, due to the mismatch in dimensionality and negativity in the embedding vectors. The transformation may cause some  loss in the knowledge and hence may make some bias for the baselines, since TPS uses the original knowledge representation. Next we would like to see the performance of those methods when directly using the original knowledge representation. In this case we have to match the dimensionality of the knowledge and the global variable in LDA.

We took LDA and three large datasets into evaluation: NYT-title, Yahoo-title, Irishtimes. All the settings are the same as in Subsection \ref{sec-unsupervised-LDA}, except that the number of topics is $K=200$ which is exactly the dimensionality of the pre-trained word embedding. To ensure non-negativity in the knowledge vectors, we normalize each embedding vector to be in $[0, 1]^{200}$.

Figure \ref{fig:LPP-orig} shows the results. We observe that the behaviors of the baselines are almost the same as in the experiments of Subsection \ref{sec-unsupervised-LDA}. One interesting thing is that KPS in this evaluation seems not to utilize the knowledge well, as its performance keeps steady or deteriorates over time. This is in contrast to the case where the knowledge is transformed into a lower dimensionality by PCA, and then input to the baselines. Figure \ref{fig:LPP-orig} suggests that TPS can utilize the knowledge well to perform significantly better than the baselines in both measures.

In summary, TPS can directly exploit an external knowledge  of different forms when learning a model, while other baselines find difficulties. TPS can use the knowledge in its original representation, while other methods often need a suitable transformation and hence do not well exploit the full strength of the external knowledge to improve a Bayesian model.

\section{Conclusion}
We presented a novel framework (TPS) that overcomes many drawbacks of existing approaches for streaming conditions. In particular, TPS exploits prior knowledge well, while other methods can forget it very fast. It has hyperparameter $\sigma$ as a simple mechanism to balance different sources of knowledge. One interesting question is how to learn $\sigma$ efficiently? 


%



\ifCLASSOPTIONcompsoc
  \section*{Acknowledgments}
\else
  \section*{Acknowledgment}
\fi

This work was funded by Gia Lam Urban Development and Investment Company Limited, Vingroup and supported by Vingroup Innovation Foundation (VINIF) under project code VINIF.2019.DA18.


\ifCLASSOPTIONcaptionsoff
  \newpage
\fi



%
\bibliographystyle{ieeetr}
\bibliography{TKDE}

\begin{thebibliography}{10}

\bibitem{blei2017VIreview}
D.~M. Blei, A.~Kucukelbir, and J.~D. McAuliffe, ``Variational inference: A
  review for statisticians,'' {\em Journal of the American Statistical
  Association}, vol.~112, no.~518, pp.~859--877, 2017.

\bibitem{Box76mispecified}
G.~E.~P. Box, ``Science and statistics,'' {\em Journal of the American
  Statistical Association}, vol.~71, no.~356, pp.~791--799, 1976.

\bibitem{huang2004applying}
Z.~Huang, H.~Chen, and D.~Zeng, ``Applying associative retrieval techniques to
  alleviate the sparsity problem in collaborative filtering,'' {\em ACM
  Transactions on Information Systems (TOIS)}, vol.~22, no.~1, pp.~116--142,
  2004.

\bibitem{hidasi2016session}
B.~Hidasi, A.~Karatzoglou, L.~Baltrunas, and D.~Tikk, ``Session-based
  recommendations with recurrent neural networks,'' in {\em International
  Conference on Learning Representations}, 2016.

\bibitem{banerjee2007clustering}
S.~Banerjee, K.~Ramanathan, and A.~Gupta, ``Clustering short texts using
  wikipedia,'' in {\em ACM SIGIR}, pp.~787--788, 2007.

\bibitem{yan2014btm}
X.~Cheng, X.~Yan, Y.~Lan, and J.~Guo, ``Btm: Topic modeling over short texts,''
  {\em IEEE Transactions on Knowledge and Data Engineering}, vol.~26, no.~12,
  pp.~2928--2941, 2014.

\bibitem{liang2017leveraging}
J.~Liang, L.~Jiang, D.~Meng, and A.~Hauptmann, ``Leveraging multi-modal prior
  knowledge for large-scale concept learning in noisy web data,'' in {\em
  Proceedings of the 2017 ACM on International Conference on Multimedia
  Retrieval}, pp.~32--40, ACM, 2017.

\bibitem{ha2019eliminating}
C.~Ha, V.-D. Tran, L.~Ngo, and K.~Than, ``Eliminating overfitting of
  probabilistic topic models on short and noisy text: The role of dropout,''
  {\em International Journal of Approximate Reasoning}, vol.~112, pp.~85--104,
  2019.

\bibitem{nguyen2021BPS}
D.~A. Nguyen, V.~L. Ngo, K.~A. Nguyen, C.~H. Nguyen, and K.~Than, ``Boosting
  prior knowledge in streaming variational bayes,'' {\em Neurocomputing},
  vol.~424, pp.~143--159, 2021.

\bibitem{tang2014understandingLDA}
J.~Tang, Z.~Meng, X.~Nguyen, Q.~Mei, and M.~Zhang, ``Understanding the limiting
  factors of topic modeling via posterior contraction analysis,'' in {\em
  Proceedings of The 31st International Conference on Machine Learning (ICML)},
  pp.~190--198, 2014.

\bibitem{yao2017incorporating}
L.~Yao, Y.~Zhang, B.~Wei, Z.~Jin, R.~Zhang, Y.~Zhang, and Q.~Chen,
  ``Incorporating knowledge graph embeddings into topic modeling.,'' in {\em
  AAAI}, pp.~3119--3126, 2017.

\bibitem{nguyen2015improving}
D.~Q. Nguyen, R.~Billingsley, L.~Du, and M.~Johnson, ``Improving topic models
  with latent feature word representations,'' {\em Transactions of the
  Association for Computational Linguistics}, vol.~3, pp.~299--313, 2015.

\bibitem{zhao2017word}
H.~Zhao, L.~Du, and W.~Buntine, ``A word embeddings informed focused topic
  model,'' in {\em Asian Conference on Machine Learning}, pp.~423--438, 2017.

\bibitem{Ideker2011knowledge}
T.~Ideker, J.~Dutkowski, and L.~Hood, ``Boosting signal-to-noise in complex
  biology: prior knowledge is power,'' {\em Cell}, vol.~144, no.~6,
  pp.~860--863, 2011.

\bibitem{andrzejewski2009incorporating}
D.~Andrzejewski, X.~Zhu, and M.~Craven, ``Incorporating domain knowledge into
  topic modeling via dirichlet forest priors,'' in {\em Proceedings of the 26th
  Annual International Conference on Machine Learning}, pp.~25--32, ACM, 2009.

\bibitem{andrzejewski2011domain}
D.~Andrzejewski, X.~Zhu, M.~Craven, and B.~Recht, ``A framework for
  incorporating general domain knowledge into latent dirichlet allocation using
  first-order logic,'' in {\em IJCAI}, vol.~22, p.~1171, 2011.

\bibitem{jagarlamudi2012lexicalpriors}
J.~Jagarlamudi, H.~Daum{\'e}~III, and R.~Udupa, ``Incorporating lexical priors
  into topic models,'' in {\em EACL}, pp.~204--213, 2012.

\bibitem{chen2013domain}
Z.~Chen, A.~Mukherjee, B.~Liu, M.~Hsu, M.~Castellanos, and R.~Ghosh,
  ``Leveraging multi-domain prior knowledge in topic models.,'' in {\em IJCAI},
  vol.~13, pp.~2071--77, 2013.

\bibitem{sweeting2008BayesianES}
M.~Sweeting, D.~De~Angelis, M.~Hickman, and A.~Ades, ``Estimating hepatitis c
  prevalence in england and wales by synthesizing evidence from multiple data
  sources. assessing data conflict and model fit,'' {\em Biostatistics},
  vol.~9, no.~4, pp.~715--734, 2008.

\bibitem{tan2018BayesianES}
S.~Tan, S.~Makela, D.~Heller, K.~Konty, S.~Balter, T.~Zheng, and J.~H. Stark,
  ``A bayesian evidence synthesis approach to estimate disease prevalence in
  hard-to-reach populations: hepatitis c in new york city,'' {\em Epidemics},
  vol.~23, pp.~96--109, 2018.

\bibitem{devlin2019bert}
J.~Devlin, M.-W. Chang, K.~Lee, and K.~Toutanova, ``Bert: Pre-training of deep
  bidirectional transformers for language understanding,'' in {\em Proceedings
  of the NAACL-HLT}, pp.~384--394, 2019.

\bibitem{Erhan2010pretrain}
D.~Erhan, Y.~Bengio, A.~Courville, P.-A. Manzagol, P.~Vincent, and S.~Bengio,
  ``Why does unsupervised pre-training help deep learning?,'' {\em Journal of
  Machine Learning Research}, vol.~11, pp.~625--660, 2010.

\bibitem{turian2010word}
J.~Turian, L.~Ratinov, and Y.~Bengio, ``Word representations: a simple and
  general method for semi-supervised learning,'' in {\em Proceedings of the
  48th Annual Meeting of the Association for Computational Linguistics}, 2010.

\bibitem{streamvb}
T.~Broderick, N.~Boyd, A.~Wibisono, A.~C. Wilson, and M.~I. Jordan, ``Streaming
  variational bayes,'' in {\em Advances in Neural Information Processing
  Systems}, pp.~1727--1735, 2013.

\bibitem{populationdis}
J.~McInerney, R.~Ranganath, and D.~M. Blei, ``The population posterior and
  bayesian modeling on streams,'' in {\em Advances in Neural Information
  Processing Systems 28}, pp.~1153--1161, 2015.

\bibitem{masegosa17power}
A.~Masegosa, T.~D. Nielsen, H.~Langseth, D.~Ramos-L{\'o}pez, A.~Salmer{\'o}n,
  and A.~L. Madsen, ``{B}ayesian models of data streams with hierarchical power
  priors,'' in {\em Proceedings of the 34th International Conference on Machine
  Learning}, vol.~70, pp.~2334--2343, PMLR, 2017.

\bibitem{faraji2018balancing}
M.~Faraji, K.~Preuschoff, and W.~Gerstner, ``Balancing new against old
  information: The role of puzzlement surprise in learning,'' {\em Neural
  computation}, vol.~30, no.~1, pp.~34--83, 2018.

\bibitem{averell2011form}
L.~Averell and A.~Heathcote, ``The form of the forgetting curve and the fate of
  memories,'' {\em Journal of Mathematical Psychology}, vol.~55, no.~1,
  pp.~25--35, 2011.

\bibitem{blei2006dynamic}
D.~M. Blei and J.~D. Lafferty, ``Dynamic topic models,'' in {\em Proceedings of
  the 23rd international conference on Machine learning}, pp.~113--120, ACM,
  2006.

\bibitem{charlin2015dynamic}
L.~Charlin, R.~Ranganath, J.~McInerney, and D.~M. Blei, ``Dynamic poisson
  factorization,'' in {\em Proceedings of the 9th ACM Conference on Recommender
  Systems}, pp.~155--162, ACM, 2015.

\bibitem{lda}
D.~M. Blei, A.~Y. Ng, and M.~I. Jordan, ``Latent dirichlet allocation,'' {\em
  Journal of Machine Learning Research}, vol.~3, pp.~993--1022, 2003.

\bibitem{lau2014machine}
J.~H. Lau, D.~Newman, and T.~Baldwin, ``Machine reading tea leaves:
  Automatically evaluating topic coherence and topic model quality,'' in {\em
  EACL}, pp.~530--539, 2014.

\bibitem{wang2006topics}
X.~Wang and A.~McCallum, ``Topics over time: a non-markov continuous-time model
  of topical trends,'' in {\em Proceedings of the 12th ACM SIGKDD International
  Conference on Knowledge Discovery and Data Mining}, pp.~424--433, ACM, 2006.

\bibitem{wei2007dynamic}
X.~Wei, J.~Sun, and X.~Wang, ``Dynamic mixture models for multiple
  time-series,'' in {\em IJCAI}, vol.~7, pp.~2909--2914, 2007.

\bibitem{wang2008cDTM}
C.~Wang, D.~Blei, and D.~Heckerman, ``Continuous time dynamic topic models,''
  in {\em Uncertainty in Artificial Intelligence}, pp.~579--586, 2008.

\bibitem{cappe2009onlineEM}
O.~Capp{\'e} and E.~Moulines, ``On-line expectation--maximization algorithm for
  latent data models,'' {\em Journal of the Royal Statistical Society: Series B
  (Statistical Methodology)}, vol.~71, no.~3, pp.~593--613, 2009.

\bibitem{Bottou18onlineLearning}
L.~Bottou, F.~E. Curtis, and J.~Nocedal, ``Optimization methods for large-scale
  machine learning,'' {\em SIAM Review}, vol.~60, no.~2, pp.~223--311, 2018.

\bibitem{doucet2001introduction}
A.~Doucet, N.~De~Freitas, and N.~Gordon, ``An introduction to sequential monte
  carlo methods,'' in {\em Sequential Monte Carlo methods in practice},
  Springer, 2001.

\bibitem{petterson2010word}
J.~Petterson, W.~Buntine, S.~M. Narayanamurthy, T.~S. Caetano, and A.~J. Smola,
  ``Word features for latent dirichlet allocation,'' in {\em Advances in Neural
  Information Processing Systems}, pp.~1921--1929, 2010.

\bibitem{xie2015incorporating}
P.~Xie, D.~Yang, and E.~Xing, ``Incorporating word correlation knowledge into
  topic modeling,'' in {\em NAACL-HLT}, pp.~725--734, 2015.

\bibitem{ibrahim2015power}
J.~G. Ibrahim, M.-H. Chen, Y.~Gwon, and F.~Chen, ``The power prior: theory and
  applications,'' {\em Statistics in medicine}, vol.~34, no.~28,
  pp.~3724--3749, 2015.

\bibitem{kpspakdd}
N.~D. Anh, N.~V. Linh, N.~K. Anh, and K.~Than, ``Keeping priors in streaming
  bayesian learning,'' in {\em Advances in Knowledge Discovery and Data Mining:
  21st Pacific-Asia Conference, Proceedings, Part II}, pp.~247--258, Springer,
  2017.

\bibitem{he2013dynamic}
Y.~He, C.~Lin, W.~Gao, and K.-F. Wong, ``Dynamic joint sentiment-topic model,''
  {\em ACM Transactions on Intelligent Systems and Technology (TIST)}, vol.~5,
  no.~1, p.~6, 2013.

\bibitem{jahnichen18a}
P.~Jähnichen, F.~Wenzel, M.~Kloft, and S.~Mandt, ``Scalable generalized
  dynamic topic models,'' in {\em Proceedings of the 21st International
  Conference on Artificial Intelligence and Statistics}, vol.~84,
  pp.~1427--1435, PMLR, 2018.

\bibitem{dieng2019dynamic}
A.~B. Dieng, F.~J. Ruiz, and D.~M. Blei, ``The dynamic embedded topic model,''
  {\em arXiv preprint arXiv:1907.05545}, 2019.

\bibitem{svi}
M.~D. Hoffman, D.~M. Blei, C.~Wang, and J.~W. Paisley, ``Stochastic variational
  inference.,'' {\em Journal of Machine Learning Research}, vol.~14, no.~1,
  pp.~1303--1347, 2013.

\bibitem{mikolov2013distributed}
T.~Mikolov, I.~Sutskever, K.~Chen, G.~S. Corrado, and J.~Dean, ``Distributed
  representations of words and phrases and their compositionality,'' in {\em
  Advances in neural information processing systems}, pp.~3111--3119, 2013.

\bibitem{bordes2013translating}
A.~Bordes, N.~Usunier, A.~Garcia-Duran, J.~Weston, and O.~Yakhnenko,
  ``Translating embeddings for modeling multi-relational data,'' in {\em
  Advances in neural information processing systems}, pp.~2787--2795, 2013.

\bibitem{parisi2019continual}
G.~I. Parisi, R.~Kemker, J.~L. Part, C.~Kanan, and S.~Wermter, ``Continual
  lifelong learning with neural networks: A review,'' {\em Neural Networks},
  vol.~113, pp.~54--71, 2019.

\bibitem{pennington2014glove}
J.~Pennington, R.~Socher, and C.~Manning, ``Glove: Global vectors for word
  representation,'' in {\em EMNLP}, pp.~1532--1543, 2014.

\bibitem{hong2010empirical}
L.~Hong and B.~D. Davison, ``Empirical study of topic modeling in twitter,'' in
  {\em Proceedings of the first workshop on social media analytics},
  pp.~80--88, ACM, 2010.

\bibitem{yan2013biterm}
X.~Yan, J.~Guo, Y.~Lan, and X.~Cheng, ``A biterm topic model for short texts,''
  in {\em Proceedings of the 22nd international conference on World Wide Web},
  pp.~1445--1456, 2013.

\end{thebibliography}



%

\vspace{-0.5in}
\newpage

\appendices

\section{Streaming Naive Bayes}
In this section, we explicitly describe the application of TPS, SVB \cite{streamvb}, and KPS \cite{kpspakdd} to \emph{multinomial Naive Bayes} (NB) for classification in document streams. It is worth noting that NB models the documents in each class $c$ by a multinomial distribution with parameter $\beta_c$. A batch learning algorithm for NB focuses mostly on estimating $\beta = (\beta_1, ..., \beta_C)$ for a classification problem with $C$ classes.\footnote{Estimating the prior for each class is  important. But for simplicity, in this study we use uniform prior over class labels.}

For TPS, the derivation is presented in the main paper. Algorithm~\ref{algo:TPS_NB} presents the streaming learning for NB by TPS.

Using variational inference, SVB \cite{streamvb}  approximates the posterior distribution of $\beta_c$ in Naive Bayes by variational distribution $q(\beta_c | \xi_c) = Dirichlet(.|\xi_c)$ where $\xi_c$ is the variational parameter associated with class $c$. Therefore learning NB is translating to learning the variational parameters $\xi_1, ..., \xi_C$. Similar to the case of LDA, we update the model at time stamp $t$ by:
\begin{align*}
\xi^t  = \xi^{t-1} + \tilde{\xi}^t 
\end{align*}
where $\xi^{t-1}$ comes from the previous minibatch $t-1$ and $\xi^0$ is initialized with  prior $\eta$.  The learned information $\tilde{\xi}^t$ from the data $D^t$ at minibatch $t$ is inferred by variational inference as
$\tilde{\xi}^t_{cv} = \sum_{d \in D_c^t} n_{dv}$, where $n_{dv}$ is the frequency of term $v$ in document $d$. Algorithm~\ref{algo:SVB_NB} summarizes the streaming learning for NB by SVB.

\begin{algorithm}[tp]
    \caption{TPS learning for Naive Bayes}
	\label{algo:TPS_NB}
\begin{algorithmic}
\REQUIRE{ Prior knowledge $\eta$, variance $\sigma$, data sequence $\{D^1,D^2,...\}$}
\ENSURE{$\pi$ \\}
Initialize $\pi^0$ randomly\\
\FOR {the $t^{th}$ minibatch} 
\STATE{Find $\pi_c^t$, for each class $c$  with dataset $D_c^t$, by using gradient ascent to maximize 
\begin{align}
LP(\pi_c^t) &= -\frac{1}{2\sigma}  \parallel \pi^t_c -\pi^{t-1}_c \parallel^2_2 \nonumber\\ 
&+ \sum_{d \in D_c^t} \sum_{n=1}^{N_d} \sum_{v=1}^V I[w_{dn}=v] {\pi_c^t} {\eta_v} \nonumber\\
&- \sum_{d \in D_c^t} \sum_{n=1}^{N_d} \sum_{v=1}^V I[w_{dn}=v] \log {\sum_{i=1}^V \exp({\pi_c^t} {\eta_i})}  \nonumber
\end{align}}
\ENDFOR
\end{algorithmic}
\end{algorithm}

\begin{algorithm}[tp]
    \caption{SVB learning for Naive Bayes}
	\label{algo:SVB_NB}
\begin{algorithmic}
\REQUIRE{ Prior knowledge $\eta$, data sequence $\{D^1,D^2,...\}$}
\ENSURE{$\xi$ \\}
Initialize $\xi^0 = \eta$\\
\FOR {The $t^{th}$ minibatch} 
\STATE{For each class $c$ with dataset $D_c^t$, compute
\begin{equation} \label{eq:NB-SVB}
\xi_{cv}^t  = \xi_{cv}^{t-1} + \sum_{d \in D_c^t} n_{dv}, \forall v
\end{equation} }
\ENDFOR
\end{algorithmic}
\end{algorithm}

KPS \cite{kpspakdd} is a variant of SVB to explicitly exploit prior knowledge $\eta$ in all minibatches. In KPS, the  model parameter $\xi^t$ in minibatch $t$ is computed as below:
\begin{align} \label{eq:NB-KPS}
\xi^t  = \xi^{t-1} + \tilde{\xi}^t + (1+t)^{-\kappa}\eta
\end{align} 
where $\kappa \ge 0$ is the dimming factor to decrease the impact of prior knowledge gradually after a number of minibatches.

\section{Qualitative evaluation on  TPS for LDA}

Interpretability is an important criteria for evaluating a model. The results from a model should be understandable and interpretable by human. In this section, we consider the interpretability/clarity of  the learned topics in LDA. In several circumstances, some methods are not able to expose a clear topic with a specific domain although that domain exists in the corpus. In this case, we choose the closest topic based on topic's keywords.  

For the evaluation on interpretability, we use two corpora: Grolier (long text) and NYT-title (short text). We fixed $K=50$ for LDA,  $\sigma = 1.0$ for TPS, $batchsize=500$ for Grolier due to its small size, and $batchsize=5000$ for NYT-title. The other settings are  the same  as those  in the Experiment part of the main paper.

Some results are shown in Table \ref{table:topwordlong} and \ref{table:topwordshort}.  While Table \ref{table:topwordlong} shows top 10 words of two topics \textit{Military} and \textit{Music} of Grolier dataset,  Table \ref{table:topwordshort} gives top words of two topics \textit{Business} and \textit{Politics} of NYT-title. The ambiguous words are written in \textit{italic} style. 

It is clear that topics learned by TPS have least ambiguous words than the other baselines. Moreover, the meaning of TPS seems to be more clear than the others with a consistent relationship of words in the topic. In addition, it is more significant for short text data than regular text. To this end, using prior knowledge is effective in term of improving the clarity of topics and making them easy to be interpreted by human.

\begin{table*}[h!]
\fontsize{8}{8} \selectfont
\caption{Top words of some learned topics of Grolier (regular text).}
\label{table:topwordlong}
\begin{center}
\begin{tabular}{lrrrrrrrrr}
\hline
\multicolumn{2}{|c|}{TPS} & \multicolumn{2}{c|}{SVB} & \multicolumn{2}{c|}{PVB} & \multicolumn{2}{c|}{KPS} & \multicolumn{2}{c|}{SVB-PP} \\ \hline
\shortstack{Topic 1 \\ (Military)} &\shortstack{Topic 2 \\ (Music)} & \shortstack{Topic 1 \\ (Military)} &\shortstack{Topic 2 \\ (Music)} & \shortstack{Topic 1 \\ (Military)} &\shortstack{Topic 2 \\ (Music)}& \shortstack{Topic 1 \\ (Military)} &\shortstack{Topic 2 \\ (Music)} &\shortstack{Topic 1 \\ (Military)} &\shortstack{Topic 2 \\ (Music)}\\ 
war      & music      & \textit{space}  & music      & war             & music         & war & music &war &music\\ 
army     & musical    & \textit{air}    & opera      & army            & musical       & \textit{king} & opera &army & opera\\ 
naval    & piano      & \textit{world}  & musical    & american        & composer      & army & musical & military & musical\\ 
navy     & songs      & soviet          & dance      & \textit{york}   & instruments   & german & piano & forces& composer\\ 
commander& composer   & flight          & ballet     & \textit{united} & \textit{century} & france & instruments &world & piano\\ 
command  & orchestral & satellite       & theater    & \textit{world}  & \textit{games}& \textit{french} & songs &naval &orchestra\\ 
military & instruments& war             & composer   & military        & songs         & \textit{germany} & composers & british & instruments\\ 
forces   & orchestra  & force           & \textit{american} & battle   & piano         & \textit{son} & composer & battle &songs\\ 
air      & vocal      & \textit{ft}     & \textit{french} & \textit{british} & player & military & operas &ship &vocal\\ 
ship     & sound      & nuclear         & \textit{stage} & forces      & composers    & battle & orchestra & aircraft & jazz\\ \hline

\end{tabular}
\end{center}
\end{table*} 

\begin{table*}
\fontsize{8}{8} \selectfont

\caption{Top words of some learned topics of NYT-title (short text).}
\label{table:topwordshort}
\begin{center}
\begin{tabular}{lrrrrrrrrr}
\hline
\multicolumn{2}{|c|}{TPS} & \multicolumn{2}{c|}{SVB} & \multicolumn{2}{c|}{PVB} & \multicolumn{2}{c|}{KPS} & \multicolumn{2}{c|}{SVB-PP}\\ \hline
\shortstack{Topic 1 \\ (Business)} &\shortstack{Topic 2 \\ (Politics)} & \shortstack{Topic 1 \\ (Business)} &\shortstack{Topic 2 \\ (Politics)} & \shortstack{Topic 1 \\ (Business)} &\shortstack{Topic 2 \\ (Politics)}& \shortstack{Topic 1 \\ (Business)} &\shortstack{Topic 2 \\ (Politics)} & \shortstack{Topic 1 \\ (Business)} &\shortstack{Topic 2 \\ (Politics)}\\ 
sell  & court   & world         & president        & sale    & obama             & dollar & obama  & buy & obama \\ 
world & vote    & europe        & election         & profit  & president         & \textit{year} & president &company & \textit{join} \\ 
plan  & obama   & profit        & reform           & run     & law               & fall & \textit{pick} &stake & debate\\ 
\textit{u.s}.  & bush           & british          & \textit{phone} & net & bush & sale & \textit{ad} &investor & ban\\ 
stock & \textit{case}& \textit{unite} & \textit{champion} & \textit{bond} & congress & million & \textit{student} & expand & \textit{challenge}\\ 
cut   & senate  & business        & threaten         & \textit{series} & \textit{press}   & market & media & \textit{challenge} & \textit{link} \\ 
buy   & clinton & \textit{chemical} & \textit{robert} & \textit{award}   &\textit{ mix} & \textit{news} & \textit{advertise} & \textit{asset} & gun\\ 
rise  & campaign & consumer     & \textit{smith}    & rise & benefit    & american & candidate& \textit{mystery} &\textit{risk} \\ 
trade & debate   & \textit{chairman} & \textit{jr.} & \textit{sea}       & missile & stock & \textit{story} &technology & island\\ 
profit& law     & \textit{magazine}& \textit{yield} & \textit{human}     & \textit{shut} & trade & \textit{event} &\textit{oversea} & \textit{spend} \\ \hline

\end{tabular}
\end{center}
\end{table*}

\section{Sensitivity of TPS with respect to parameters}
In this section,  we investigate the effects of the parameters:  number $K$ of topics, batchsize, and  variance $\sigma$. Both regular text (Grolier) and short text (TMN-title) are used in our evaluation.

%

\subsection{Sensitivity of TPS for LDA with respect to the number of topics}
\begin{figure}[tp]
\centering
\includegraphics[width=0.5\textwidth]{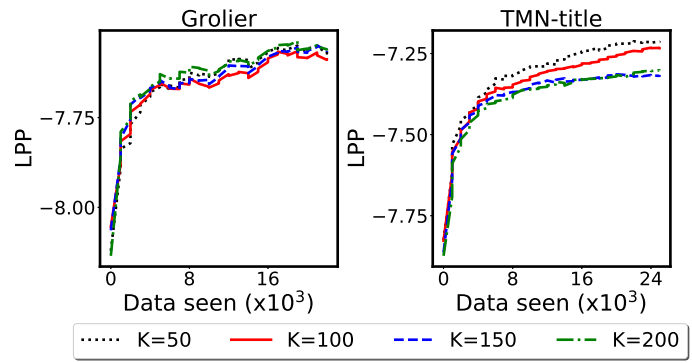}
\caption{Sensitivity of TPS with respect to the number $K$ of topics when LDA is the base model. \label{fig:k}}
\end{figure}

We fix batchsize = 500,  $\sigma=1$, and the number of topics is tested in $[50,100,150,200]$.  The results are shown in Figure \ref{fig:k} for  regular text (Grolier) and short text (TMN-title) respectively. While TPS is stable in regular text when changing $K$, it seems to be more sensitive with short text than regular text. Moreover, the smaller number of topics can make TPS do well in short text.      

\subsection{Sensitivity of TPS  for LDA with respect to batchsize}
\begin{figure}[tp]
\centering
\includegraphics[width=0.5\textwidth]{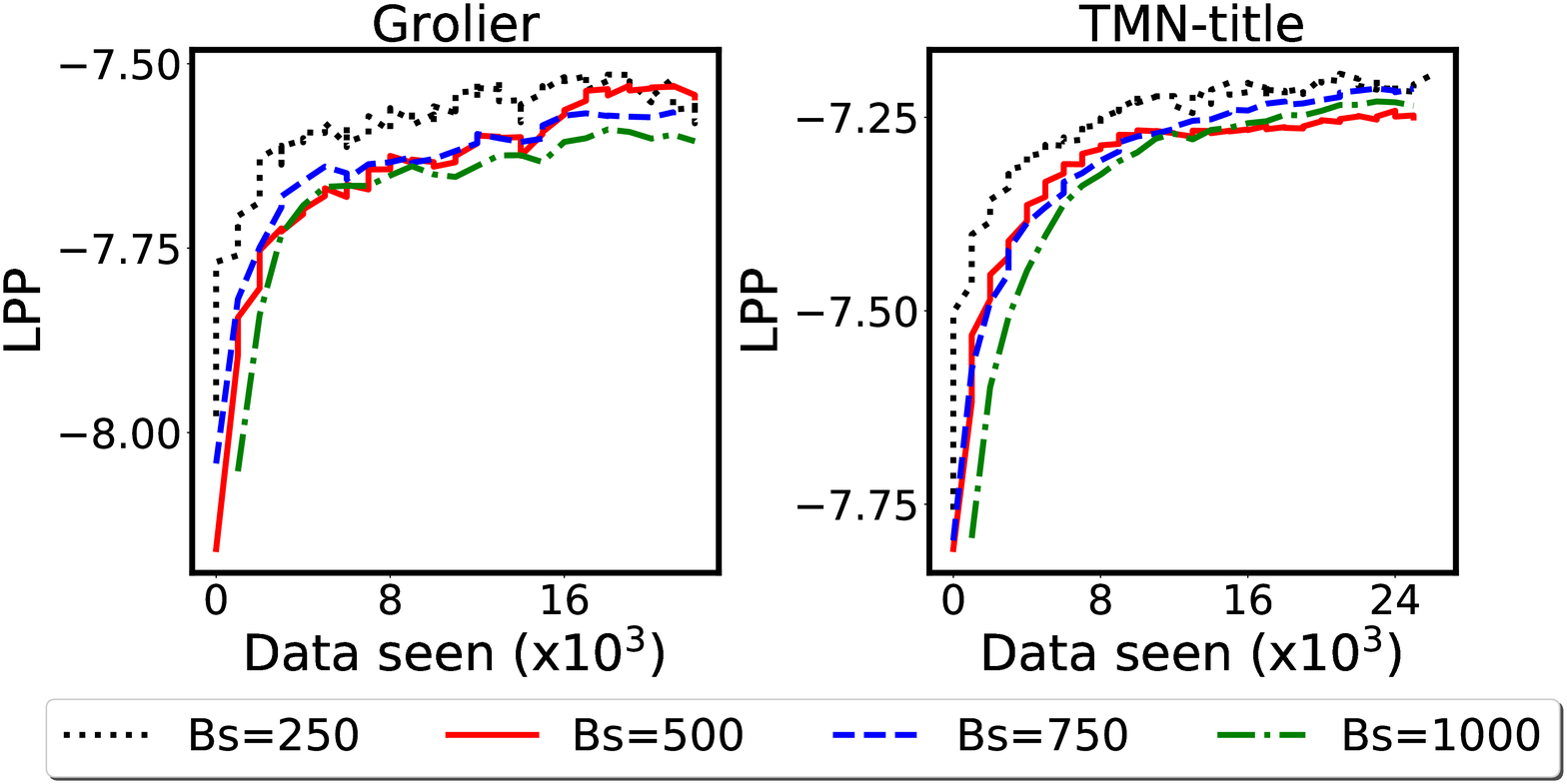} 
\caption{Sensitivity of TPS with respect to batchsize when LDA is the base model. \label{fig:batch}}
\end{figure}

To examine the sensitivity of TPS over batchsize, we fix $\sigma=1.0, K=50$.  The results are shown in Figure  \ref{fig:batch}. We can see that batchsize has some similar impact on regular and short text. From the assumption in TPS, the streaming data is processed in each data collection decided by batchsize which means this parameter determines the information from new arrived data to balance with prior knowledge and the past minibatch. Therefore, TPS seems to be more sensitive on batchsize than $K$.    

\subsection{Sensitivity of TPS with respect to the variance in Naive Bayes}
\begin{figure}[!t]
\centering
\includegraphics[width=0.5\textwidth]{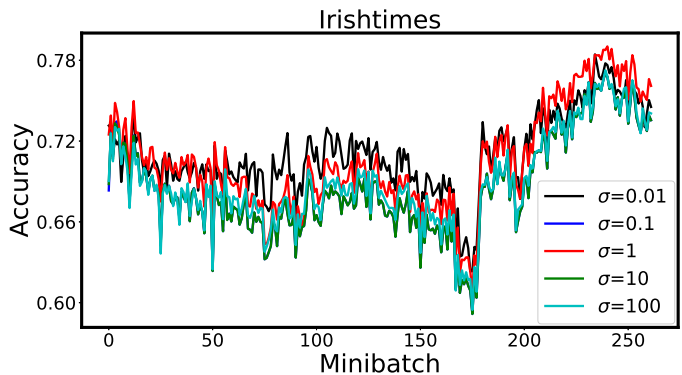}
\caption{Sensitivity of TPS for Naive Bayes w.r.t the variance $\sigma$. \label{fig:NBsigma_classification}}
\end{figure}

Figure \ref{fig:NBsigma_classification} shows the sensitivity of TPS w.r.t $\sigma$. It seems that large $\sigma (\ge 10)$ seem to perform worse than smaller values of $\sigma$. $\sigma \le 1$ seems to be good, meaning that the model at each minibatch should not be  far from that in the previous minibatch. The accuracy gap among the settings is noticeable in a number of the first minibatches. However, the difference gradually decreases as more data arrive.

\section{Multi-pass training for TPS}
We examine the effectiveness of prior and transition of TPS in multi-pass training, which passes (iterates) the whole training data more than one time.

\begin{figure}[!t]
\centering
\includegraphics[width=0.45\textwidth]{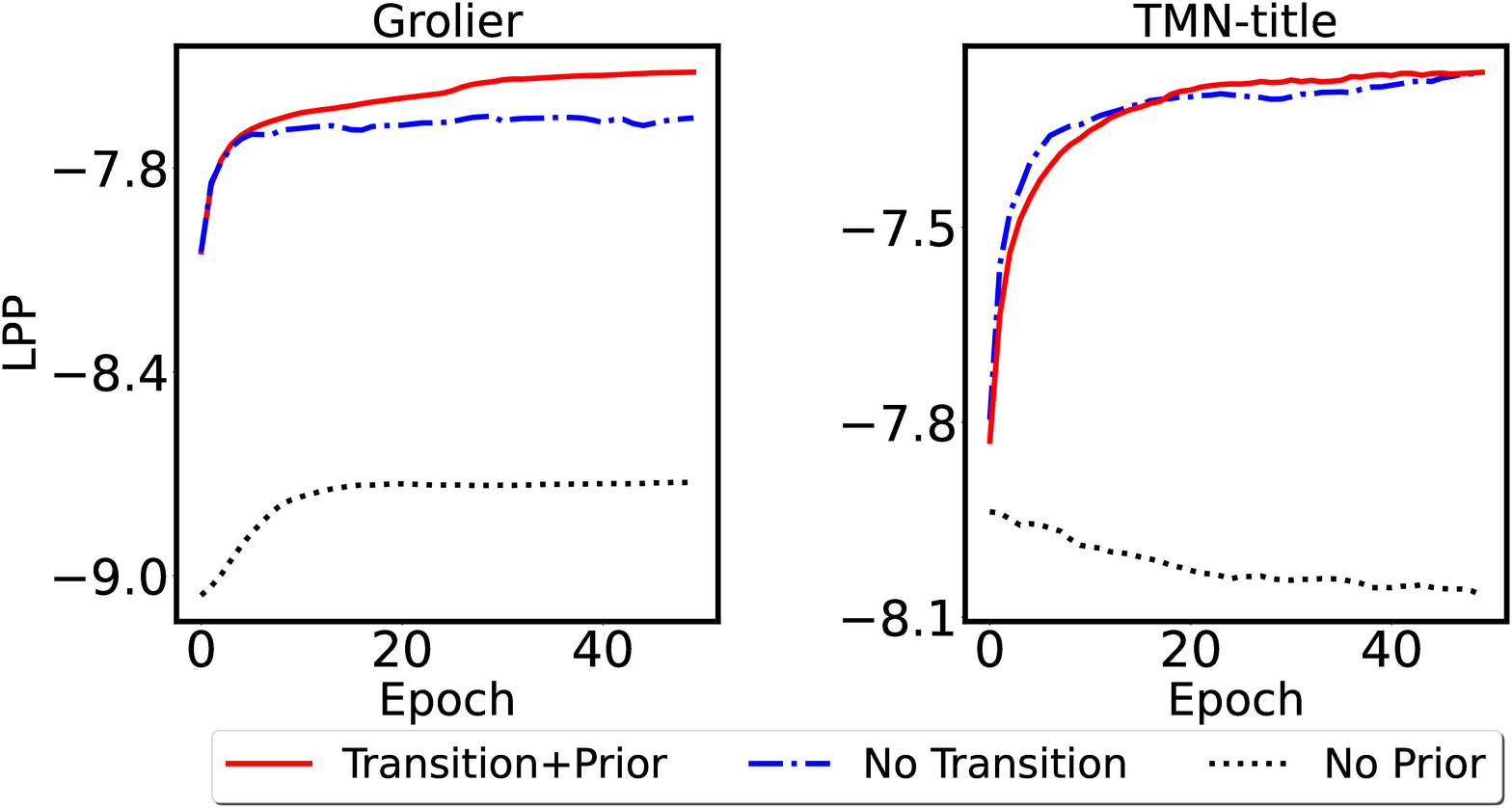}
\caption{The effect of each components (Transition, Prior) of TPS in multi-pass training, i.e., multiple passes over the whole data are allowed when learning. LDA is the base model. \label{fig_off:multipass}}
\end{figure}

In detail, we pass through the data 50 times, each time is an epoch. After each epoch, we evaluate the log predictive probability of the model. The results for Grolier and TMN-title are shown in Fig. \ref{fig_off:multipass}.


We again see that having Prior (for Transition + Prior and No transition) is significantly better than No Prior. This again confirms the importance of using prior knowledge. 

We also see that Transition + Prior and No transition have comparable performances. We can explain as that multi-pass training pass through the data many times, hence the model can still learn the transformation of prior knowledge without using transition.

Moreover, multi-pass training allows TPS achieving higher performance in LPP than single pass training.  However, in reality, the infinite and large incoming data prevent us from passing data multiple times. We need to balance the trade off between the performance and the resources of storage and running time.

\section{Running Time}

\begin{figure}[!t]
\centering
\includegraphics[width=0.45\textwidth]{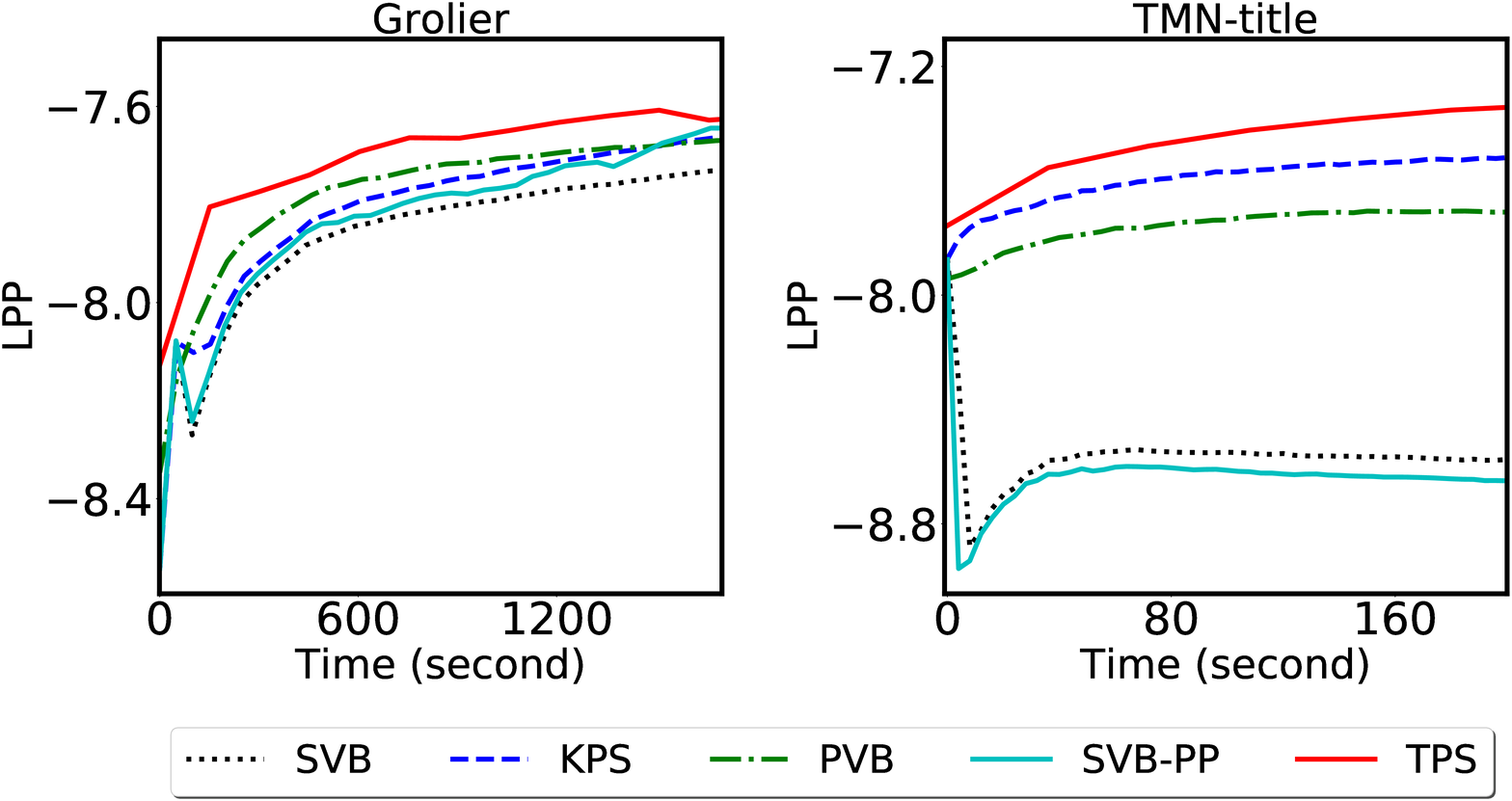}
\caption{Performance of five methods as  learning time increases.  \textit{Pre-trained word embeddings} is the prior knowledge and LDA is the base model. \label{fig:times}}
\end{figure}

We measured the running time of all methods with two datasets (Grolier and TMN-tile) similar to sensitivity analysis. We fix $\sigma=1$, $k=50$, and $batch size=500$ for Grolier and TMN-title. The results are shown in Table \ref{table-time} and Figure~\ref{fig:times}. 

We observe that TPS can obtain a high performance at a faster rate than other baselines. This is surprising. One reason may be that  the inference for each document in the baselines takes significant times due to the computation of the expectation of Dirichlet distribution for global parameters.
Meanwhile TPS does not need to compute such an expectation. Therefore, TPS runs faster than the baselines.

\begin{table}
\caption{Running time (in seconds) of all methods.}
\label{table-time}
\begin{center}
\begin{tabular}{lrrrrr}
\hline
Datasets & SVB  & SVP-PP  & PVB  & KPS & TPS \\ \hline
Grolier   & 2260 & 2272 & 2358 & 2383 & 1741 \\ 
TMN-title & 244 & 250 & 309 & 252 & 240 \\ \hline
\end{tabular}
\end{center}
\end{table}

\section{Details of the evaluation metrics}
\subsection{Log predictive probability}
We follow the metric used in \cite{svi}. Generally, given the model learned from training data $D$, each document in the evaluation set is divided into two disjoint parts: the held-out words $w_{ho}$ and observed words $w_{obs}$. The local variables are inferred using $w_{obs}$, then the predictive probability of the model is evaluated by the log probability:
\begin{align*}
\log p(w_{ho} | D, w_{obs})
\end{align*}
In a $LDA$ model with $K$ topics, the global word topic distributions $\beta$, and the document-specific distribution $\theta$, we have: 

\begin{align*}
&p(w_{ho} | D, w_{obs})\\
&= \int \int (\sum_{1}^{K} \theta_k \beta_{k,w_{ho}}) p(\theta | w_{obs},\beta) p(\beta | D) d\theta d\beta 
\end{align*}
\begin{align*}\\
                    &\approx \int \int (\sum_{1}^{K} \theta_k\beta_{k,w_{ho}}) q(\theta) q(\beta) d\theta d\beta \\
                    &= \sum_{k=1}^{K} E_q[\theta_k]E_q[\beta_{k,w_{ho}}]
\end{align*}
where $q(\beta)$ and $q(\theta)$ are approximate distribution of variables $\beta$ and $\theta$ respectively. Note that when $\beta$ is point estimation, $E_q[\beta]$ is replaced by $\beta$, and $\theta$ is inferred from observed words $w_{obs}$ given $\beta$.

\subsection{Normalized pointwise mutual information (NPMI)}
This metric was introduced by \cite{lau2014machine}. NPMI score give an evaluation for correlation with human-judged coherence. In detail, given a topic $t$ with top-$T$ topic words $w_1, w_2,...,w_T$, the $NPMI$ score for topic $t$ is calculated by:
\begin{align*}
NPMI(t) = \sum_{1 \leq i < j \leq N } {\frac{\log \frac{P(w_i,w_j)}{P(w_i)P(w_j)}}{-\log P(w_i,w_j)}} 
\end{align*}
where $P(w_i)$ is the probability of word $w_i$ derived from corpus and $P(w_i,w_j)$ is the probability of co-occurrence of two words $w_i$ and $w_j$ in the same document.

\end{document}